\documentclass[letterpaper]{article} 
\usepackage{aaai2026}  
\usepackage{times}  
\usepackage{helvet}  
\usepackage{courier}  
\usepackage[hyphens]{url}  
\usepackage{graphicx} 
\usepackage{natbib}  
\usepackage{caption} 
\usepackage{algorithm}
\usepackage{algorithmic}
\usepackage{multirow}
\usepackage{multicol}
\usepackage{color}
\usepackage{todonotes}
\usepackage{amsmath}
\usepackage{tabularx}
\usepackage{color, colortbl}
\usepackage{threeparttable}
\usepackage{enumitem}
\usepackage{booktabs}

\usepackage{newfloat}
\usepackage{listings}
\DeclareCaptionStyle{ruled}{labelfont=normalfont,labelsep=colon,strut=off} 
\floatstyle{ruled}
\newfloat{listing}{tb}{lst}{}
\floatname{listing}{Listing}
\title{Labels or Input? Rethinking Augmentation in Multimodal Hate Detection}
\author{
   Sahajpreet Singh,
    Kokil Jaidka,
    Subhayan Mukerjee
}
\affiliations{
    National University of Singapore\\
    \texttt{jaidka@nus.edu.sg}
}

\usepackage{bibentry}

\nocopyright

\begin{document}

\maketitle

\begingroup
\renewcommand\thefootnote{}\footnotetext{\textcolor{red}{\textbf{Warning:} This paper includes instances of hateful memes and offensive language. These don’t reflect the authors’ views.}}
\addtocounter{footnote}{-1}
\endgroup

\begin{abstract}
Online hate remains a significant societal challenge, especially as multimodal content enables subtle, culturally grounded, and implicit forms of harm. Hateful memes embed hostility through text–image interactions and humor, making them difficult for automated systems to interpret. Although recent Vision-Language Models (VLMs) perform well on explicit cases, their deployment is limited by high inference costs and persistent failures on nuanced content. This work examines how far small models can be improved through prompt optimization, fine-tuning, and automated data augmentation. We introduce an end-to-end pipeline that varies prompt structure, label granularity, and training modality, showing that structured prompts and scaled supervision significantly strengthen compact VLMs. We also develop a multimodal augmentation framework that generates counterfactually neutral memes via a coordinated LLM–VLM setup, reducing spurious correlations and improving the detection of implicit hate. Ablation studies quantify the contribution of each component, demonstrating that prompt design, granular labels, and targeted augmentation collectively narrow the gap between small and large models. The results offer a practical path toward more robust and deployable multimodal hate-detection systems without relying on costly large-model inference.
\end{abstract}

\begin{links}
    \link{Code}{https://github.com/sahajps/Labels-or-Input}
    \link{Dataset}{https://hf.co/datasets/sahajps/Meme-Sanity}
\end{links}

\section{Introduction}
The rise of Vision-Language Models (VLMs) such as CLIP \cite{radford2021learning}, Flamingo \cite{alayrac2022flamingo}, and InternVL2 \cite{chen2024internvl} has led to significant progress in understanding and processing multimodal content. These models are increasingly applied to downstream tasks, including meme classification and online content moderation. At the same time, visual harmful content has surged across platforms like X and Instagram, where images and videos constitute between 30\% of shared content on text-centric platforms and nearly 100\% on image-first platforms \cite{peng2023agenda, yang2023visual, pfeffer2023just}. This shift underscores the growing need for scalable, robust methods that can effectively address the complexities of multimodal hate speech detection in contemporary social media environments \cite{gonzalez2023populist, heley2022missing, solea2023mainstreaming}. Moreover, recent studies find that VLMs are more susceptible to harmful content than their unimodal counterparts, with image–text combinations inducing unsafe behaviors even under benign conditions \cite{hee2025demystifying}.

A central concern in computational social science is the quality of data used to study online harm. High-quality data depends on high-quality annotation, yet annotation cannot be carried out arbitrarily: inconsistent or weak supervision introduces artifacts, reinforces cultural biases, and destabilizes moderation systems \cite{lee2024exploring, masud2024hate}. This creates a methodological gap, particularly acute for multimodal content, between the complexity of online harm and the limited supervision that existing datasets can reliably provide.

While prior work has addressed aspects of hate meme detection, many existing approaches focus on fine-tuning pre-trained models without exploring how prompting might be used to enrich supervision or guide classification \cite{lippe2020multimodal, hermida2023detecting}. Although prompt engineering has shown promise in unimodal tasks as a lightweight method for influencing model behavior \cite{furniturewala2024thinking}, its utility in multimodal settings, especially for generating higher-quality labels or shaping classification decisions in VLMs, remains underexplored. These research gaps motivate our core research question:
\textbf{\textit{Is improving prompting or refining the underlying data more effective than a standard baseline in multimodal hateful content classification?}}

\begin{figure*}[!t]
\centering
\includegraphics[width=\textwidth]{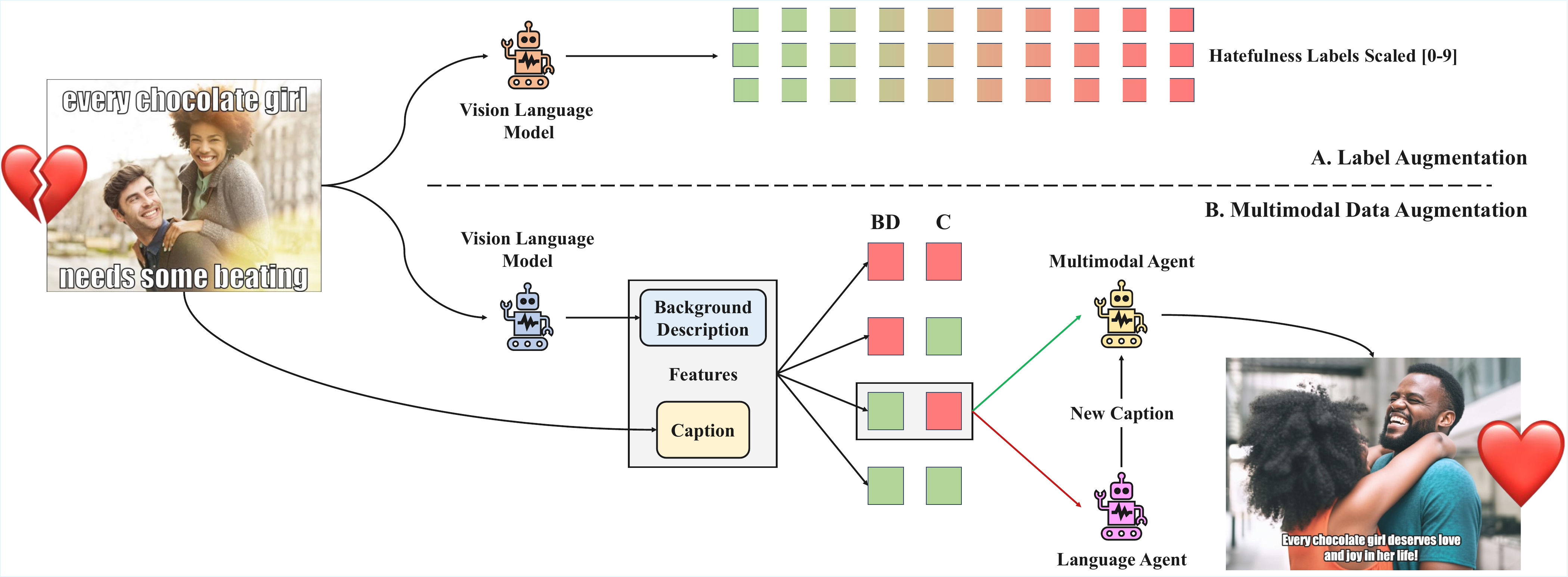}
\caption{Overview of the augmentation pipeline. Red and green boxes indicate hateful and non-hateful content. In part B, the augmented data is created for the green–red pairs—cases where the background is non-hateful but is used in combination with a hateful caption, resulting in a hateful interpretation.}
\label{fig:into_pipeline}
\end{figure*}

Recent studies also highlight significant limitations in dataset quality and data augmentation practices. Several datasets are constructed through manual pipelines that involve overlaying text on stock images, yielding templated meme formats \cite{nandi2024safe}. Broader surveys on multimodal hate content \cite{hee2024recent} emphasize the growing presence of hateful speech across text, image, and video, but often stop short of proposing actionable modeling strategies. Synthetic augmentation methods such as MixGen \cite{hao2023mixgen} attempt to increase diversity by randomly mixing images and captions, but frequently produce semantically incoherent pairings that reduce realism. In parallel, cultural and representational biases embedded in VLMs \cite{madasu2025cultural, liu2025culturevlm} are increasingly recognized yet receive limited attention in multimodal hate speech detection, particularly for culturally grounded hateful memes \cite{bui2025multi3hate}.

In this study, we investigate two complementary strategies (see Figure \ref{fig:into_pipeline}) for improving hateful meme classification: (1) \textbf{prompt optimization}, which leverages structured prompting to enrich supervision across modalities and label formats; and (2) \textbf{multimodal augmentation}, which generates counterfactually neutral memes to reduce spurious correlations. These approaches offer alternatives to enhance dataset quality, mitigate bias, and conduct a thorough multimodal analysis of social media content.

We conduct experiments on the Facebook Hateful Memes dataset \cite{kiela2020hateful}, which contains over 10K image–caption pairs labeled for binary hatefulness. Our evaluation involves fine-tuning both unimodal and multimodal models under each strategy. This setup enables a direct comparison of prompt-informed supervision and data-centric augmentation with standard baselines for robust multimodal classification. Our study finds that label augmentation, such as scaling hatefulness, shows consistent patterns across modalities and prompting styles. This indicates that well-augmented multimodal hate labels can enable a more nuanced understanding of harmful content beyond simple binary classification. We also observe that input-level multimodal augmentation improves performance across modalities when fine-tuning pretrained models. Human evaluation further supports the quality of the augmented data, highlighting its potential in reducing toxicity and pointing to the broader promise of large multimodal models when used strategically in multi-agent systems.

\subsubsection{Contributions} In line with our ethical commitment, we deliberately avoid generating new hateful content in the name of multimodal data augmentation. The prevalence of such material on social media is already deeply concerning, and our goal is to mitigate harm, not amplify it. In short, this work makes the following key contributions:
\begin{itemize}
\item We extend the Facebook Hateful Memes dataset by generating \textbf{2479 counterfactually neutral memes}, offering a novel benchmark for evaluating robustness to spurious text–image correlations.
\item We introduce a structured \textbf{prompt optimization framework} for multimodal classification, demonstrating that prompt design, long studied in language-only settings, has overlooked yet significant effects in vision–language tasks.
\item We explore \textbf{multimodal augmentation as a debiasing strategy}, providing a proof-of-concept pipeline that could generalize to related domains such as cultural bias mitigation in VLMs.
\end{itemize}

\section{Related Work}
Humor and satire are frequently used to mask derogatory content or reinforce stereotypes while maintaining plausible deniability \cite{smuts2010ethics}. As noted in prior work, this form of disparagement humor can reduce social sanctions for expressing prejudice and foster political intolerance \cite{perks2012superiority,ferguson2008disparagement, masood2024intolerance}, provoking reactive sharing behaviors that reinforce in-group superiority and out-group hostility \cite{masood2024intolerance}. These dynamics make hatefulness in memes both culturally contingent and pragmatically difficult to detect through surface features alone.

To address the difficulty of interpreting memes, prior work has shifted from unimodal to multimodal approaches. Early efforts relied on isolated text or image classifiers, but these often failed to capture the semantic interplay that drives hateful meaning in memes \cite{lippe2020multimodal}. The release of the Hateful Memes dataset \cite{kiela2020hateful} catalyzed the development of multimodal fusion models capable of integrating visual and linguistic features.

Subsequent models incorporated pre-trained vision-language architectures such as CLIP \cite{radford2021learning} and Flamingo \cite{alayrac2022flamingo}, enabling improved joint reasoning over text and image inputs. However, many of these systems still struggle with out-of-distribution examples and culturally specific memes, which are underrepresented in benchmark datasets \cite{bui2025multi3hate, madasu2025cultural}. Vision-language models exhibit known cultural and representational biases, which can be especially harmful in hate detection contexts. These biases stem from skewed pretraining distributions and insufficient coverage of minority identities or cultural idioms \cite{liu2025culturevlm, madasu2025cultural}. Additionally, VLMs are more prone than unimodal models to unsafe behaviors triggered by multimodal inputs \cite{liu2024safety}.

Such vulnerabilities highlight the need for both architectural and data-centric solutions. Our contributions involving prompt optimization and semantically coherent augmentation function as lightweight forms of regularization that improve interpretability and mitigate failure modes related to spurious correlations, stereotype reinforcement, and brittle visual-textual associations.

In summary, the need for improved multimodal hate speech detection has motivated two complementary research directions: using prompting to better structure supervision and labels, and using input data augmentation to improve robustness.

\subsubsection{Prompting as a mechanism for supervision} While prompting has become an effective tool for adapting large language models \cite{furniturewala2024thinking}, its applications in multimodal settings are less explored \cite{wu2024visual}. Existing VLMs rarely leverage structured prompts to guide classification decisions or clarify task framing. Yet, the presence or absence of hateful intent often hinges on subtle shifts in phrasing or cultural markers. Our approach introduces task- and label-specific prompt templates to serve as soft supervision, helping models differentiate between hateful and non-hateful combinations of text and image, particularly in ambiguous or sarcastic cases.

\subsubsection{Augmentation as a tool for improving robustness} In parallel, prior work highlights the limitations of hate meme datasets built from templated or synthetic formats \cite{nandi2024safe}, and the failure of augmentation techniques like MixGen \cite{hao2023mixgen} to preserve semantic coherence. On the other hand, some studies \cite{koushik2025camu, gu2025mememind} use the textual direction of data augmentation by adding extra context (such as a detailed description of the image, CoT reasoning process, etc.). While converting inputs to a single modality (such as text) can improve hate tagging, this approach is not particularly effective for debiasing memes during fine-tuning. The training process may still cause the model to associate certain labels with specific image component words like ``Muslim" or ``Black". To address this, we adopt a counterfactual data augmentation strategy that constructs minimally modified meme variants, e.g., replacing hateful captions with neutral ones while retaining visual features. This helps models disentangle surface form from semantic content, reducing overfitting to visual cues and improving generalization to culturally grounded memes.

\section{Methodology}
We evaluate two distinct strategies for improving hate meme classification: (1) \textbf{prompt optimization}, which treats prompting as a multivariate design space across learning modality, structure, label format, and fine-tuning; and (2) \textbf{multimodal augmentation}, which seeks to improve model robustness by systematically neutralizing hateful captions while preserving visual context. These approaches are compared to assess their effectiveness in enhancing classification accuracy, robustness, and fairness.

\subsection{Prompt Optimization Framework}
For our first set of experiments, we explore a full-factorial design spanning 3 learning styles (text-only: fine-tuning, multimodal: fine-tuning \& prompting), 2 prompt strategies (simple vs. category-based), and 2 output types (binary vs. scale-based). We use prompting as a labeling instrument to generate enriched supervision signals. Specifically, we design structured prompt templates ranging from simple to category-specific prompts and from binary to scale-based outputs, and use them with a teacher model (GPT-4o-mini\footnote{\url{https://platform.openai.com/docs/models/gpt-4o-mini}}) on training data to support scale-based labeling and generate interval annotations (0–9), thereby enhancing label granularity. These enhanced labels are then used to fine-tune both a large multimodal model (InternVL2 4B \cite{chen2024internvl}) and a smaller unimodal model (BERT \cite{devlin2019bert}, RoBERTa \cite{liu2019roberta}), allowing us to evaluate the impact of prompt-informed supervision. 

\subsubsection{Problem formulation}
Let each input meme be defined as a tuple $(I_i, T_i, y_i)$, where $I_i$ is the image, $T_i$ is the caption text, and $y_i \in \{0, 1\}$ is the binary ground-truth hatefulness label. Given a model $M$ parameterized by $\theta$, a prompt template $P$, and a label format $L$, the model's predicted label $\hat{y}_i$ is generated as:

\[
\hat{y}_i = M(I_i, T_i; \theta, P, L)
\]

\subsubsection{Prompt Construction} We define two strategies:
\begin{itemize}
    \item \textbf{Simple Prompt ($P_s$)}: A direct instruction that asks whether the meme is hateful, without further elaboration.
    \item \textbf{Category Prompt ($P_c$)}: A structured prompt that defines subtypes of hate (e.g., misogyny, xenophobia, political hate), encouraging the model to attend to specific cues. 
\end{itemize}

All prompt templates are modular and composable with both label formats (see details in Appendix). We fix prompt wording and structure across conditions to minimize confounds.

\subsubsection{Label Format Design}

Two types of outputs are evaluated:
\begin{itemize}
    \item \textbf{Binary Output ($L_b$)}: The model is learning to tag hatefulness label as \texttt{TRUE} or \texttt{FALSE}.
    \item \textbf{Scale Output ($L_s$)}: The model is getting trained on a hatefulness score on an integer scale from 0 (not hateful) to 9 (extremely hateful). And later, at the test time, it is being mapped to \texttt{TRUE} (5-9) or \texttt{FALSE} (0-4).
\end{itemize}

To enable fine-tuning on scale outputs, we use a teacher model to generate soft labels for training. Incorrect labels (i.e., those inconsistent with the binary ground truth) are filtered out to ensure label quality.

Ultimately, this results in a $3 \times 2 \times 2$ factorial experiment across learning modality ($M$), prompt structure ($P$), and label format ($L$) with 12 distinct configurations. Each configuration is evaluated independently to isolate the marginal contributions of each component and to test for interaction effects between prompting and fine-tuning.

\subsection{Multimodal Augmentation}
The Facebook Hateful Memes dataset exhibits several known artifacts, including templated image-text combinations, culturally skewed representations, and a heavy concentration of caption-driven hatefulness. These are the structural limitations that motivate our second set of experiments, involving counterfactual augmentation strategy.

We developed a multi-step and multi-agent pipeline to regenerate non-hateful variants of memes using LLM and VLM tools. This approach produces a bias-aware augmented dataset designed to reduce spurious correlations while preserving semantic and visual coherence.

\subsubsection{Problem Formulation}

Let $D = \{(I_i, T_i, y_i)\}_{i=1}^N$ denote a dataset of memes, where $I_i$ is the image, $T_i = [w_1, \ldots, w_{|T_i|}]$ is the tokenized caption text, and $y_i \in \{0, 1\}$ is the binary hatefulness label. We define a counterfactual data augmentation objective to construct neutral variants $\tilde{M}_i = (\tilde I_i, \tilde{T}_i, \tilde{y}_i)$ by replacing hateful captions $T_i$ with context-preserving neutral captions $\tilde{T}_i$, where $\tilde{y}_i = 0$.

This augmentation targets memes for which the caption is the primary hateful modality, i.e., where $I_i$ is semantically benign. The hypothesis is that injecting such counterfactuals will decorrelate textual toxicity from visual features and improve generalization.

\subsubsection{Hatefulness Attribution}

To identify the modality responsible for hatefulness, we define a modality-level attribution function:
\[
H_i = f_\phi(g_\psi(I_i), T_i) \in \{\texttt{image}, \texttt{text}, \texttt{both}, \texttt{none}\}
\]

Where $f_\phi$ is a zero-shot classifier instantiated via Qwen2.5-14B \cite{qwen2.5} and $g_\psi$ is defined in the forthcoming paragraph. Only samples with $H_i = \texttt{text}$ are selected for augmentation. This procedure ensures that alterations are made only to memes where the caption is the dominant hateful component.

\subsubsection{Context-Preserving Caption Neutralization}

We first generate a background description $D_i$ for the image $I_i$ using a vision-language model $g_\psi$ (InternLM-XComposer-2.5-7B \cite{internlmxcomposer2_5}): $D_i = g_\psi(I_i)$

The model is constrained to exclude any recognition of overlaid text. Given the original caption $T_i$ and image context $D_i$, we generate a rewritten, neutral caption $\tilde{T}_i$ using a generative model $h_\omega$ (GPT-4o-mini):
\[
\tilde{T}_i = h_\omega(T_i,D_i) 
\text{, such that } \texttt{Hate}(\tilde{T}_i) = 0
\]

We enforce semantic coherence and topical grounding through conditioning on $D_i$. While $h_\omega$ includes built-in safety mechanisms to promote safe responses, we still assess its safety performance as part of our human evaluation on a subset of the augmented multimodal dataset.

\subsubsection{Meme Regeneration and Dataset Augmentation}

We synthesize a regenerated meme $\tilde{M}_i = (\tilde I_i, \tilde{T}_i, 0)$ using a rendering function $r_\gamma$ implemented via Gemini 2.0 Flash (experimental)\footnote{\url{https://cloud.google.com/vertex-ai/generative-ai/docs/models/gemini/2-0-flash}}, where, $\tilde I_i = r_\gamma (D_i,\tilde T_i)$. This model overlays $\tilde{T}_i$ on $I_i$'s equivalent using layout-preserving visual information. The final augmented dataset is defined as:
\[
\tilde{D} = \{ \tilde{M}_i = (\tilde I_i, \tilde{T}_i, 0) \mid H_i = \texttt{text}, \texttt{Hate}(\tilde{T}_i) = 0 \}
\]
Where $g_\psi(\tilde I_i) \approx g_\psi(I_i)$, i.e., the background descriptions of old and newly augmented memes are similar.

This corpus $\tilde{D}$ is merged with the original dataset $D$ for downstream training. We evaluate its impact on accuracy, fairness, and out-of-distribution robustness. Table \ref{tab:ex_mm_aug} shows some visual examples for the pipeline.

\subsubsection{Meme Typology and Target Class}
We define four meme types based on modality-specific hatefulness: HH (hateful image \& hateful caption), HN (hateful image \& non-hateful caption), NH (non-hateful image \& hateful caption), and NN (non-hateful image \& caption).

Empirical analysis indicates that the majority of hateful memes belong to the \textbf{NH} category. Our augmentation strategy explicitly targets this class to reduce spurious associations between benign visual features and toxic labels.

Full implementation details, including model prompts and architecture-specific settings, are provided in the Appendix.

\section{Empirical Evaluation}
We now describe the datasets, baselines, and models used to evaluate our two experimental approaches, prompt optimization and multimodal augmentation, for the task of classifying hate memes. Our evaluation focuses on classification accuracy and weighted-F1 score, as standard in the hate classification benchmarks. 

\begin{table*}[!t]
    \centering
    \resizebox{\textwidth}{!}{
    \begin{tabular}{|l|c|c|c|c|}
        \hline
        & \textbf{Example 1} & \textbf{Example 2} & \textbf{Example 3} & \textbf{Example 4} \\
        \hline \hline
        
        \rotatebox[origin=l]{90}{\textbf{\small Examples}}
        & \includegraphics[height=0.25\textwidth]{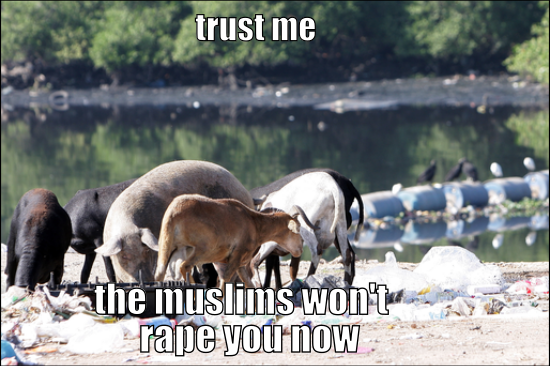}
        & \includegraphics[height=0.25\textwidth]{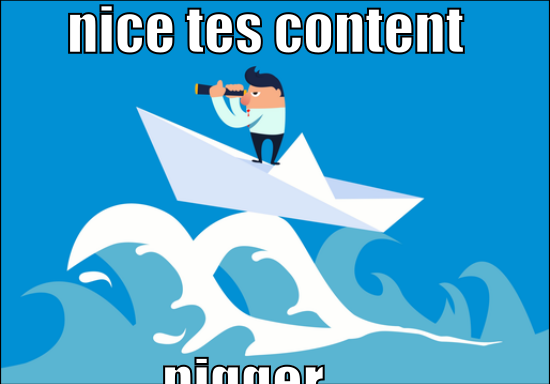}
        & \includegraphics[height=0.25\textwidth]{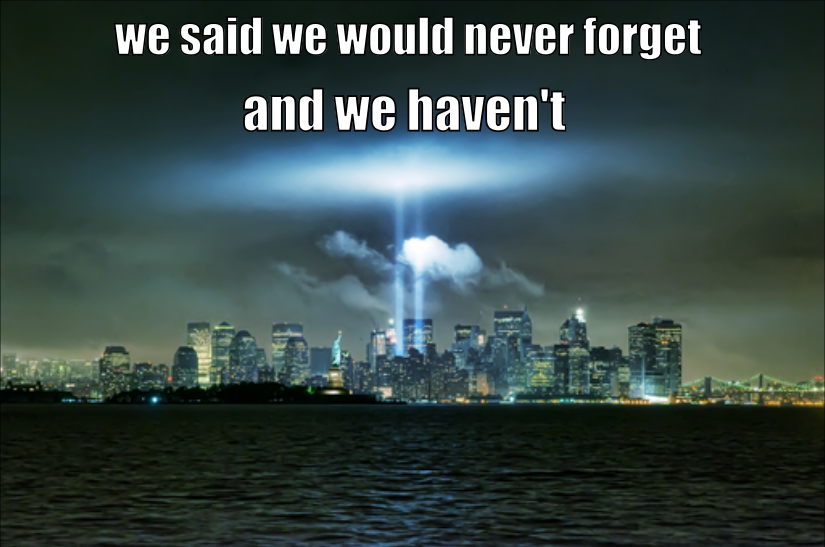}
        & \includegraphics[height=0.25\textwidth]{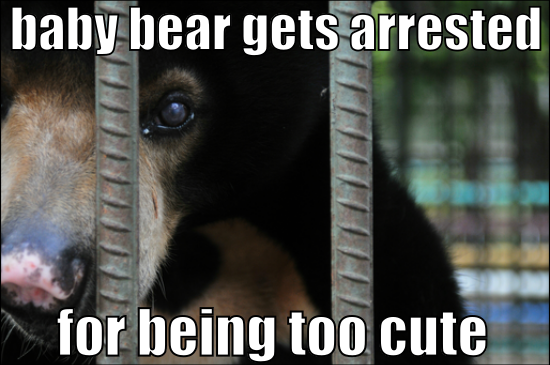} \\ \hline

        \rotatebox[origin=l]{90}{\textbf{\small HS*}} & 9 & 9  & 0 & 0 \\ \hline
    \end{tabular}}
    \caption{Consistent examples of the scaled labels.}
    \label{tab:ex_label_aug}
\end{table*}

\begin{table*}[!t]
    \centering
    \resizebox{\textwidth}{!}{
    \begin{tabular}{|l|c|c|c|c|}
        \hline
        & \textbf{Example 1} & \textbf{Example 2} & \textbf{Example 3} & \textbf{Example 4} \\
        \hline \hline
        
        \rotatebox[origin=l]{90}{\textbf{\small Hateful Meme}}
        & \includegraphics[height=0.25\textwidth]{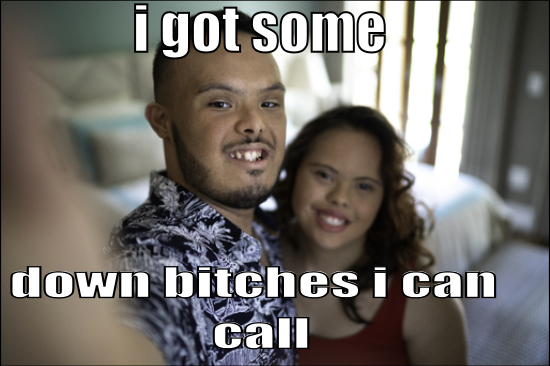}
        & \includegraphics[height=0.25\textwidth]{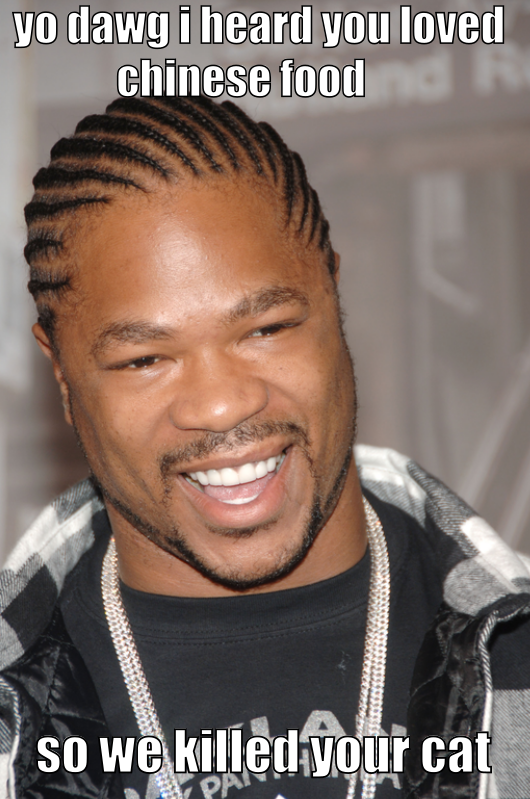}
        & \includegraphics[height=0.25\textwidth]{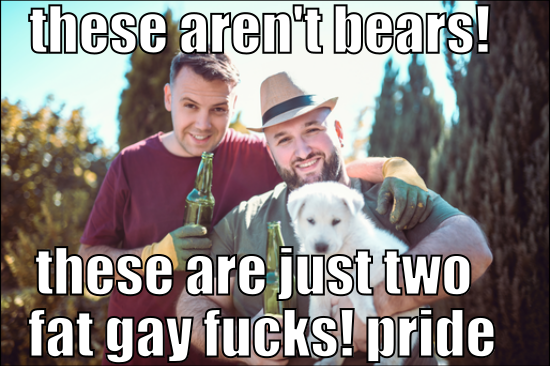}
        & \includegraphics[height=0.25\textwidth]{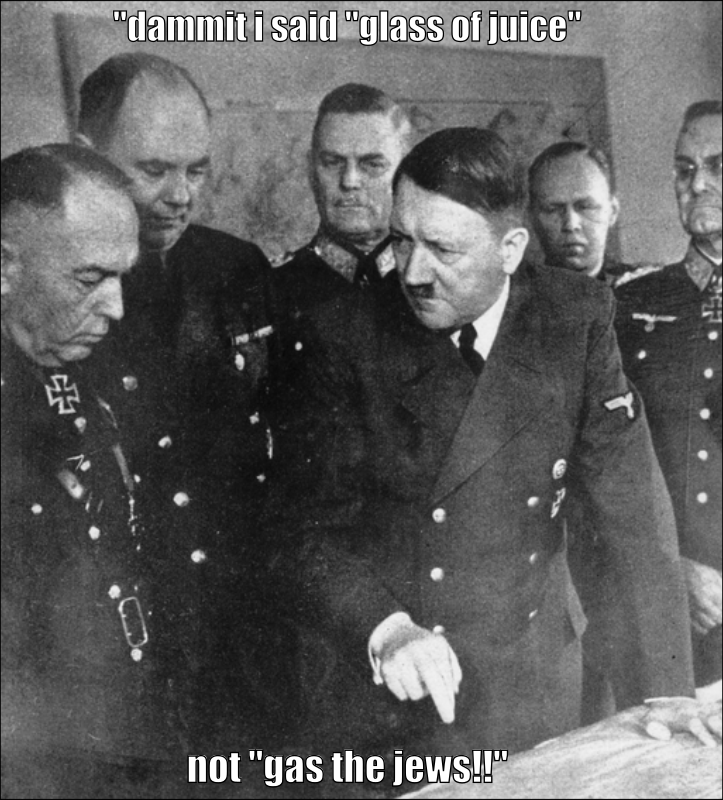} \\ \hline
        
        \rotatebox[origin=l]{90}{\textbf{\small Non-hateful Augmentation}}
        & \includegraphics[height=0.25\textwidth]{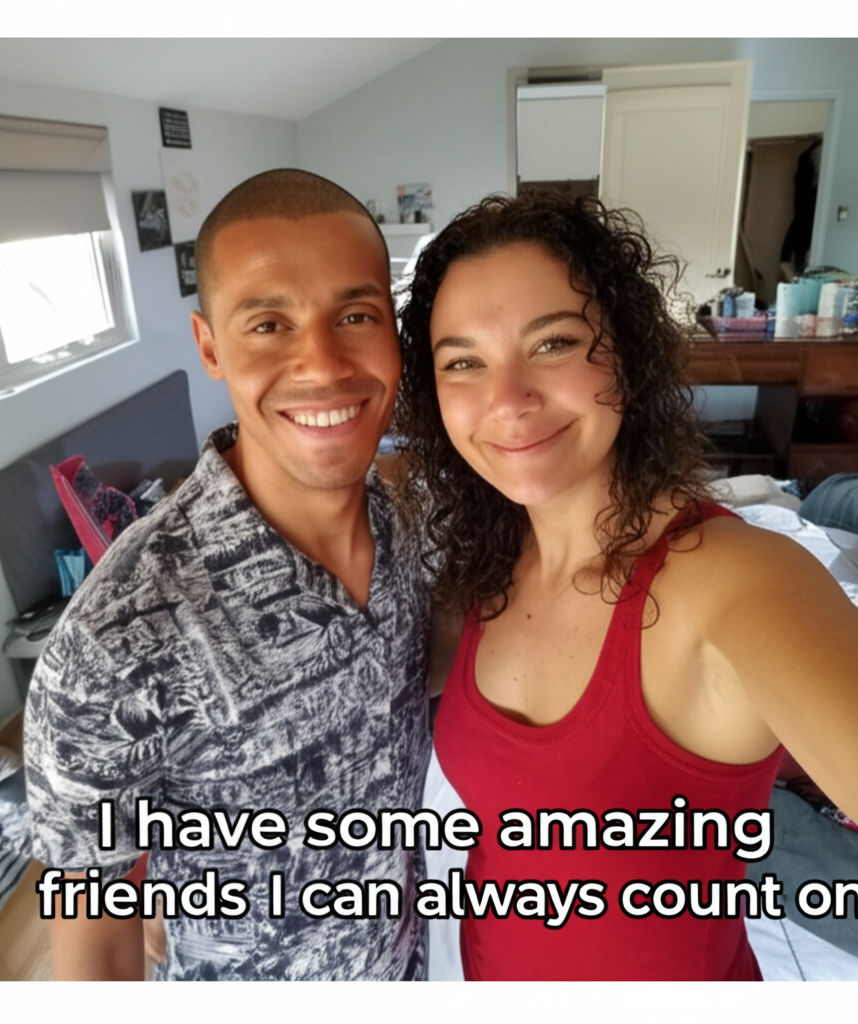}
        & \includegraphics[height=0.25\textwidth]{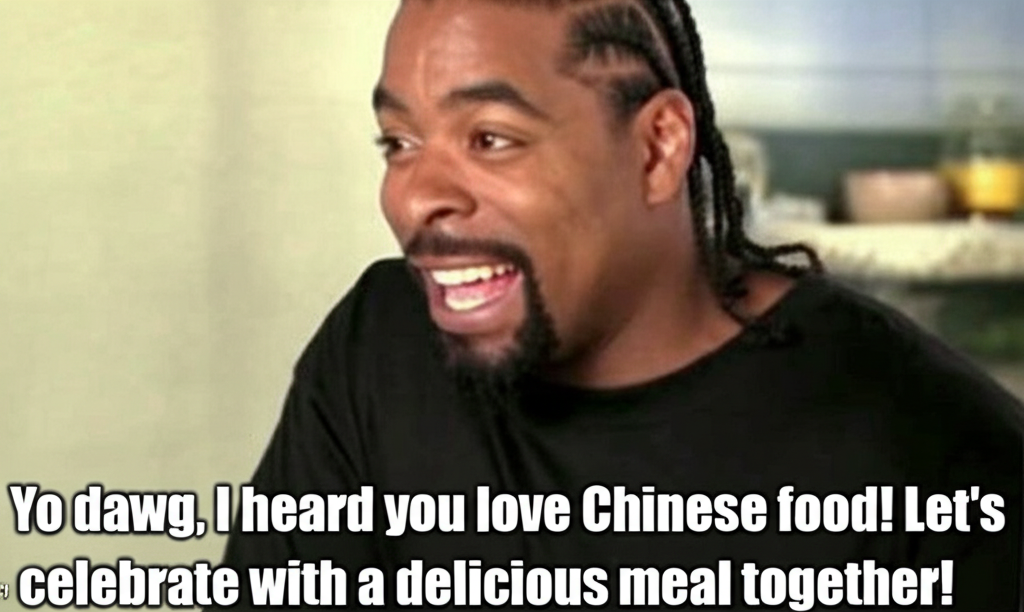}
        & \includegraphics[height=0.25\textwidth]{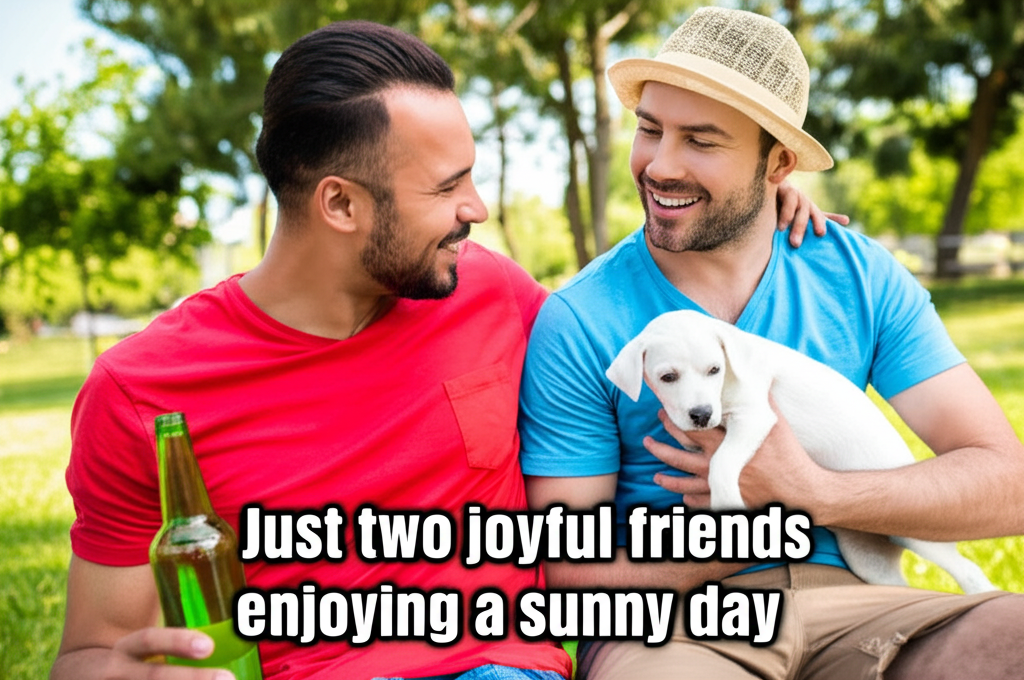}
        & \includegraphics[height=0.25\textwidth]{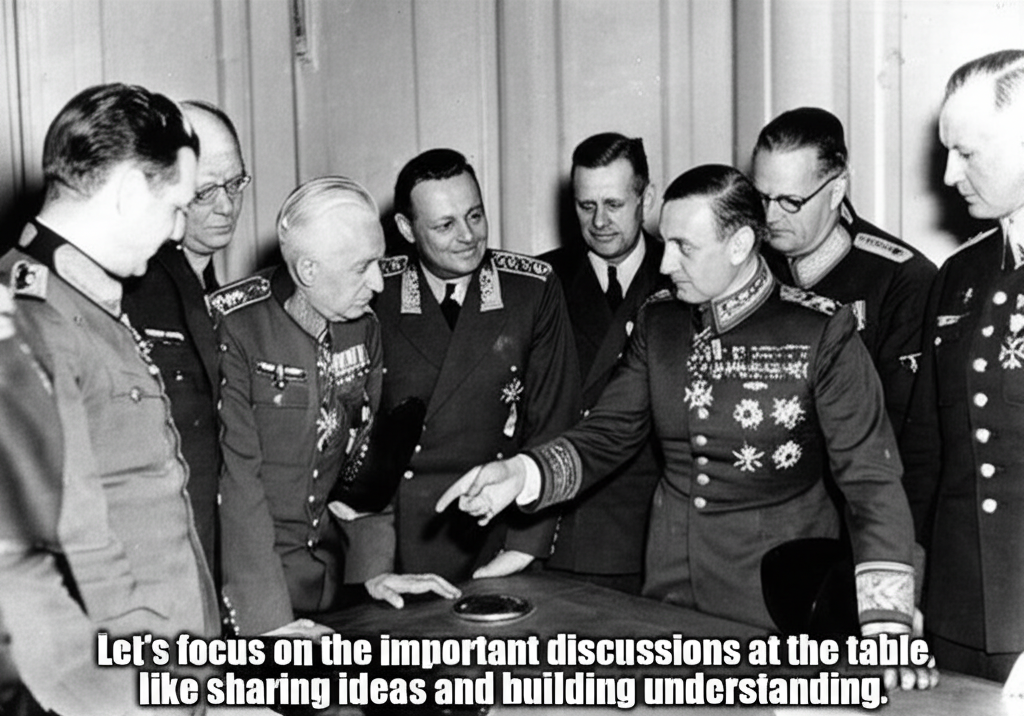} \\
        \hline \hline
        
        & \textbf{Example 5} & \textbf{Example 6} & \textbf{Example 7} & \textbf{Example 8} \\
        \hline \hline
        
        \rotatebox[origin=l]{90}{\textbf{\small Hateful Meme}}
        & \includegraphics[height=0.25\textwidth]{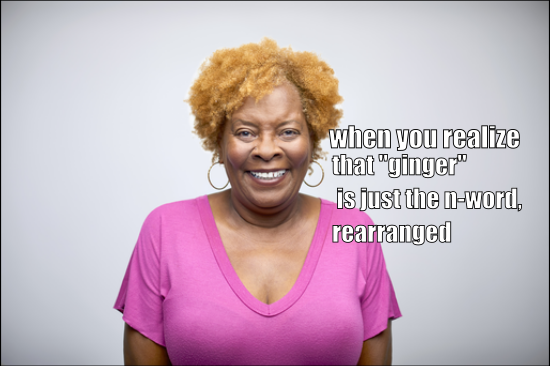}
        & \includegraphics[height=0.25\textwidth]{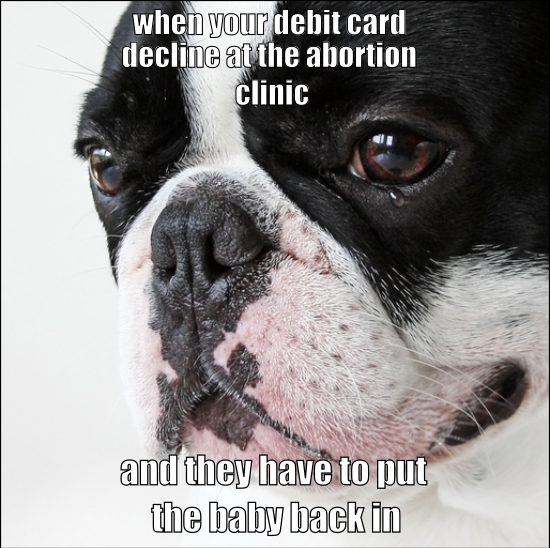}
        & \includegraphics[height=0.25\textwidth]{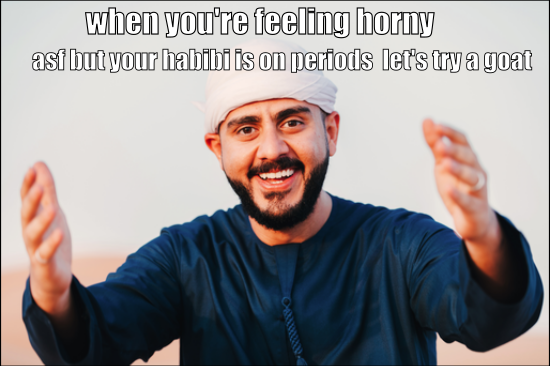}
        & \includegraphics[height=0.25\textwidth]{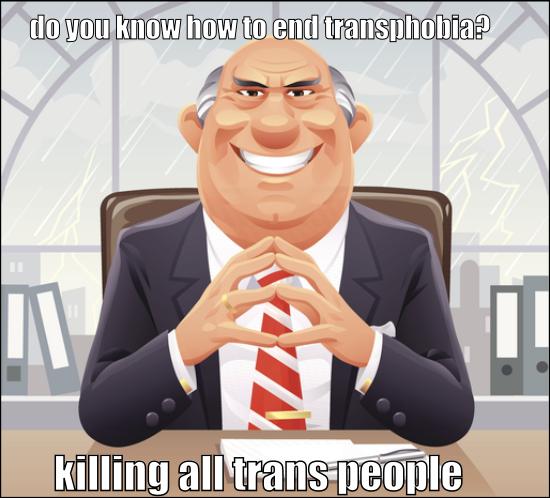} \\ \hline
        
        \rotatebox[origin=l]{90}{\textbf{\small Non-hateful Augmentation}}
        & \includegraphics[height=0.25\textwidth]{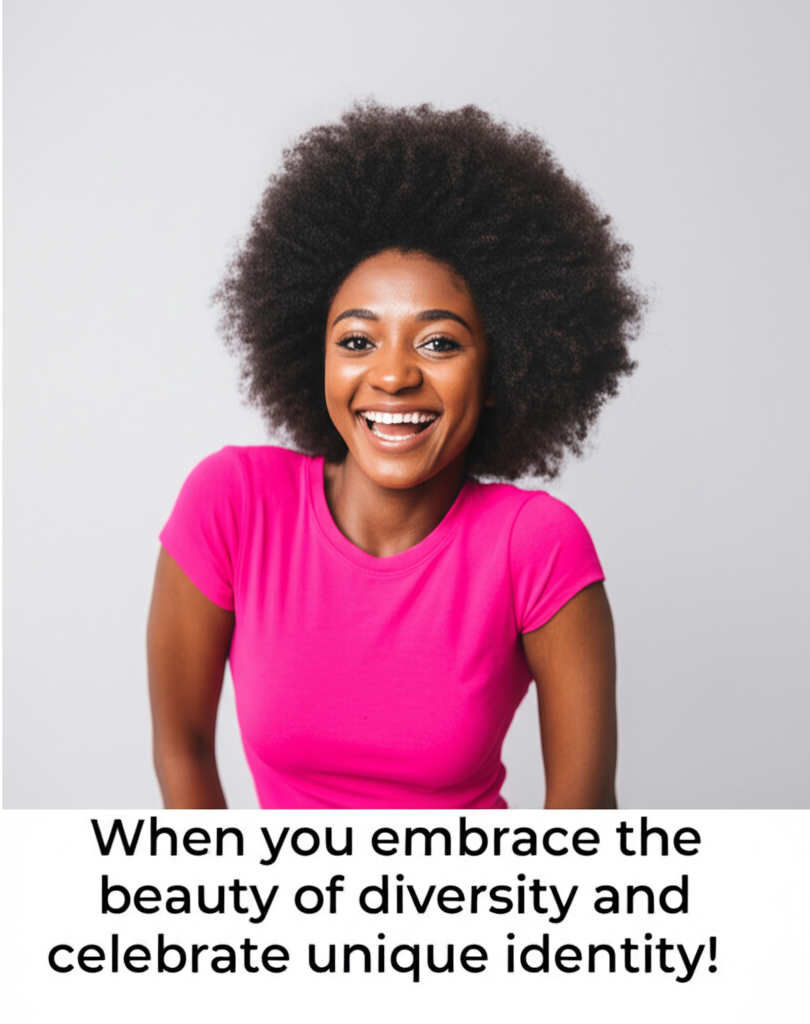}
        & \includegraphics[height=0.25\textwidth]{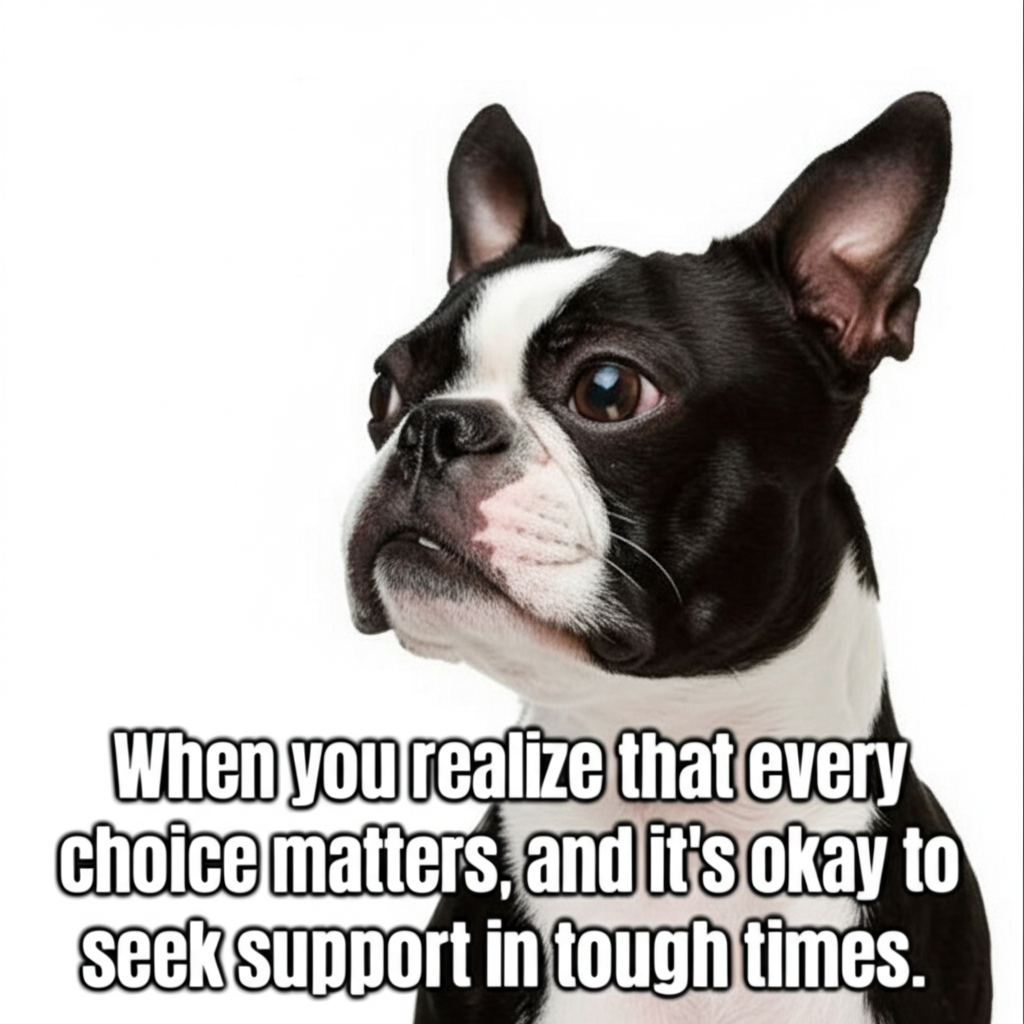}
        & \includegraphics[height=0.25\textwidth]{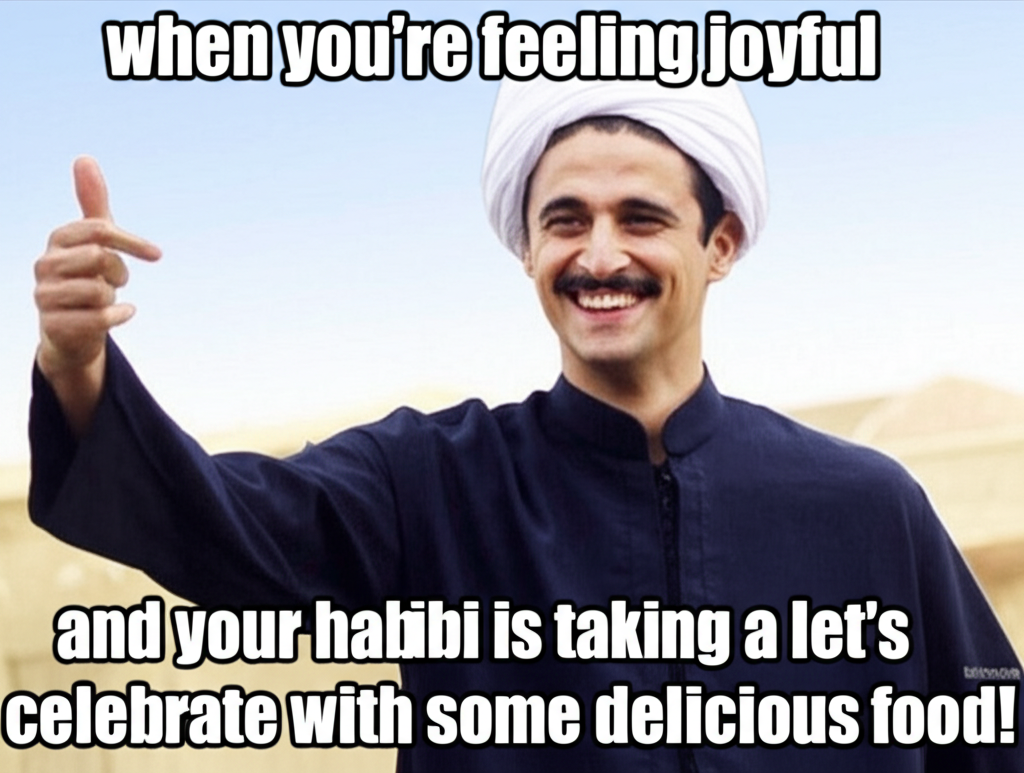}
        & \includegraphics[height=0.25\textwidth]{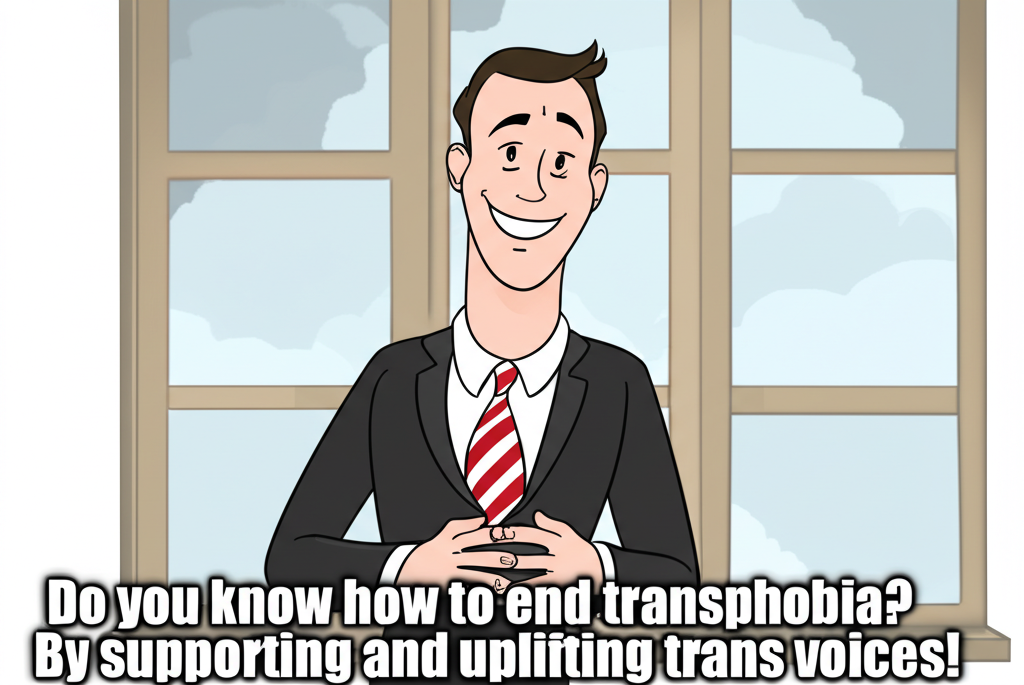} \\
        \hline
    \end{tabular}}
    \caption{Input-output examples from multimodal data augmentation pipeline.}
    \label{tab:ex_mm_aug}
\end{table*}

\subsection{Datasets}

We use the Facebook Hateful Memes dataset~\cite{kiela2020hateful}, a benchmark corpus for multimodal hate speech detection. The dataset consists of 12.9K image-caption pairs for training, 1.04K for validation, and 3K for test evaluation. Each meme is annotated with a binary label indicating whether it is hateful. Captions are written in natural language and paired with images that range from benign stock photos to visually suggestive content.

For the prompt optimization experiments, we use the original dataset labels (binary) and augment the training and validation sets with scale-based annotations, followed by misaligned example removal. By doing this, we end up with approximately 12K and 1K examples for training and validation, respectively.

For the multimodal augmentation experiments, we extract a subset of the training set in which hatefulness is localized to the caption only (i.e., image visuals are neutral). These samples are regenerated using our multi-step and multi-agent pipeline to create counterfactually non-hateful memes while preserving visual context. Using our multi-agent pipeline, we generate 2,479 counterfactually neutral variants, creating an extended dataset. Unlike the prompt optimization track, we do not use the original validation split. Instead, both configurations are balanced via random downsampling and split 80:20 into new train-validation sets, yielding approximately 9.56K/2.39K (Original) and 11.06K/2.77K (Extended) for training/validation, respectively. This ensures controlled distributions in both configurations while allowing an extended set to benefit from additional augmented data.

\subsection{Fine-Tuning Models and Setup}

To evaluate both label augmentation and multimodal data augmentation strategies, we fine-tune (sometimes prompt, where applicable, such as InternVL2) a suite of unimodal and multimodal models on the Facebook Hateful Memes dataset. For fine-tuning, AdamW is used as an optimizer with cross-entropy as a loss function, which is standard in classification tasks. All training was conducted on commodity GPUs (8×NVIDIA H100 GPU cluster) using consistent hyperparameters across models. Models used are publicly available checkpoints or APIs. Hyperparameters are also kept consistent across settings to control for confounding effects. Training and inference configurations are included in the Appendix.

\subsubsection{(a) Prompt Optimization Experiments}

For the prompt optimization track, we evaluate 12 experimental configurations from a $3 \times 2 \times 2$ factorial design varying:

\begin{itemize}
    \item \textbf{Learning Type}: using text (fine-tuning), multimodal (fine-tuning and prompting);
    \item \textbf{Prompt Style}: simple vs. category-based;
    \item \textbf{Label Format}: binary vs. scale-based labels for learning.
\end{itemize}

We compare performance across zero-shot prompting with InternVL2 and prompt-informed fine-tuning, using InternVL2 and BERT-based models (BERT, RoBERTa). All smaller models are fully fine-tuned with the same hyperparameters to ensure comparability. For InternVL2, we apply LoRA \cite{hu2022lora}-based parameter-efficient tuning. Results and model sizes are given in Table \ref{tab:label_aug_results}.

\subsubsection{(b) Multimodal Augmentation Experiments}
For this augmentation track, we fine-tune each pre-trained model on both the original dataset and an extended version, which includes counterfactually neutral memes. We experiment with three types of encoder-based models: unimodal (text), unimodal (vision), and multimodal (text + vision), allowing a direct comparison of robustness to spurious correlations. We also test the current state-of-the-art small VLMs in zero-shot and fine-tuning settings. 
For encoder-based models, fine-tuning is performed on the last two layers of the model, and for VLMs' fine-tuning, we apply QLoRA \cite{dettmers2023qlora}-based parameter-efficient tuning. Tables \ref{tab:mm_aug_results} and \ref{tab:vlm_zs} present the accuracy and F1-scores for all models.

\begin{table*}[!t]
\centering
\resizebox{0.78\textwidth}{!}{
\begin{tabular}{llclcclcc}
\hline
    \multirow{3}{*}{\textbf{Category}} & \multirow{3}{*}{\textbf{Model}} & \multirow{3}{*}{\textbf{Size}} & \multirow{3}{*}{\textbf{Prompt}} & \multicolumn{5}{c}{\textbf{Training dataset}} \\ \cline{5-9} 
    &  &  &  & \multicolumn{2}{c}{\textbf{Original (Binary)}} &  & \multicolumn{2}{c}{\textbf{Augmented (Scaled)}} \\ \cline{5-6} \cline{8-9} 
    &  &  &  & Accuracy & \multicolumn{1}{c}{F1-score} &  & Accuracy & F1-score \\ \hline
 
    \multirow{4}{*}{\begin{tabular}[c]{@{}l@{}}Unimodal\\ (fine-tuning)\end{tabular}} & \multirow{2}{*}{BERT} & \multirow{2}{*}{112M} & Simple & 60.53 $\pm$ 1.78 & 58.88 $\pm$ 1.87 &  & 61.01 $\pm$ 1.70 & 57.30 $\pm$ 1.86 \\
    &  &  & Category & 54.11 $\pm$ 1.99 & 53.41 $\pm$ 2.08 & & 60.75 $\pm$ 1.85 & 59.18 $\pm$ 1.96 \\ \cline{2-9} 
 
    & \multirow{2}{*}{RoBERTa} & \multirow{2}{*}{125M} & Simple & 61.54 $\pm$ 1.75 & 59.13 $\pm$ 1.86  & \multicolumn{1}{c}{}  &  60.79 $\pm$ 1.95 & 57.42 $\pm$ 2.08 \\
    & & & Category &  53.55 $\pm$ 1.91& 52.87 $\pm$ 2.01&  & 61.49 $\pm$ 1.85 & 58.24 $\pm$ 2.09 \\ \hline
 
    \multirow{2}{*}{\begin{tabular}[c]{@{}l@{}}Multimodal\\ (prompting)\end{tabular}} 
    & \multirow{2}{*}{InternVL2} & \multirow{2}{*}{4B} & Simple  & 61.63 $\pm$ 1.85 & 54.51 $\pm$ 2.33 & & 63.49 $\pm$ 2.02  & 60.00 $\pm$ 2.25\\
    &  &  & Category &  61.81 $\pm$ 1.93 &53.11 $\pm$ 2.43 &  & 64.46 $\pm$ 1.91 & 64.74 $\pm$ 1.89 \\ \hline
    \multirow{2}{*}{\begin{tabular}[c]{@{}l@{}}Multimodal\\ (fine-tuning)\end{tabular}} & \multirow{2}{*}{InternVL2} & \multirow{2}{*}{4B} & Simple & 68.15 $\pm$ 1.77 & 67.28 $\pm$ 1.83 &  & 64.98 $\pm$ 1.79& 63.06 $\pm$ 1.93 \\
    &  &  & Category & 67.97 $\pm$ 1.82 &  67.04 $\pm$ 1.89&  & 65.30 $\pm$ 1.76 & 63.28 $\pm$ 1.92 \\ \hline
\end{tabular}
}
\caption{Classification results with binary and scaled labels on simple and category styled prompts, in which the scales were true-only generations taught by GPT-4o-mini.
}
\label{tab:label_aug_results}
\end{table*}

\begin{table*}[!t]
\centering
\resizebox{0.78\textwidth}{!}{
\begin{tabular}{llccccccc}
\hline
\multirow{3}{*}{\textbf{Category}} & \multirow{3}{*}{\textbf{Model}} & \multirow{3}{*}{\textbf{Size}} & \multirow{3}{*}{\textbf{Learning}} & \multicolumn{5}{c}{\textbf{Training dataset}} \\ \cline{5-9} 
 &  &  &  & \multicolumn{2}{c}{\textbf{Original}} &  & \multicolumn{2}{c}{\textbf{Augmented (Extended)}} \\ \cline{5-6} \cline{8-9} 
 &  &  &  & Accuracy & F1-score &  & Accuracy & F1-score \\ \hline
\multirow{3}{*}{\begin{tabular}[c]{@{}l@{}}Unimodal\\ (fine-tuning)\end{tabular}} & ALBERT & 12M & \multirow{3}{*}{Text} & 58.48$\pm$1.87 & 54.96$\pm$2.10 &  & 58.39$\pm$1.90 & 58.54$\pm$1.91 \\
 & BERT & 112M &  & 59.08$\pm$1.95 & 55.50$\pm$2.26 &  & 58.20$\pm$2.05 & 57.95$\pm$2.08 \\
 & RoBERTa & 125M &  & 58.94$\pm$1.96 & 47.18$\pm$2.40 &  & 57.10$\pm$1.96 & 52.54$\pm$2.16 \\ \hline
\multirow{3}{*}{\begin{tabular}[c]{@{}l@{}}Unimodal\\ (fine-tuning)\end{tabular}} & ViT & 86M & \multirow{3}{*}{Vision} & 56.56$\pm$1.84 & 53.28$\pm$2.01 &  & 58.56$\pm$2.14 & 53.58$\pm$2.46 \\
 & BEiT & 86M &  & 57.87$\pm$2.08 & 53.38$\pm$2.39 &  & 58.56$\pm$2.25 & 55.43$\pm$2.46 \\
 & DeiT & 86M &  & 56.01$\pm$2.29 & 52.58$\pm$2.45 &  & 57.72$\pm$2.00 & 51.99$\pm$2.26 \\ \hline
\multirow{4}{*}{\begin{tabular}[c]{@{}l@{}}Multimodal\\ (fine-tuning)\end{tabular}} & CLIP & 150M & \multirow{4}{*}{Both} & 58.34$\pm$1.88 & 47.71$\pm$2.37 &  & 58.56$\pm$2.06 & 51.77$\pm$2.41 \\
 & SmolVLM & 2B &  & \multicolumn{1}{l}{59.65$\pm$2.03} & \multicolumn{1}{l}{51.12$\pm$2.51} & \multicolumn{1}{l}{} & \multicolumn{1}{l}{59.91$\pm$2.13} & \multicolumn{1}{l}{54.53$\pm$2.47} \\
 & Qwen2.5-VL & 3B &  & \multicolumn{1}{l}{64.16$\pm$1.85} & \multicolumn{1}{l}{60.08$\pm$2.14} & \multicolumn{1}{l}{} & \multicolumn{1}{l}{64.82$\pm$1.85} & \multicolumn{1}{l}{62.01$\pm$2.10} \\
 & Qwen2-VL & 7B &  & \multicolumn{1}{l}{70.76$\pm$1.72} & \multicolumn{1}{l}{69.59$\pm$1.85} & \multicolumn{1}{l}{} & \multicolumn{1}{l}{71.85$\pm$1.70} & \multicolumn{1}{l}{71.31$\pm$1.77} \\ \hline
\end{tabular}
}
\caption{Classification results with and without newly augmented non-hateful meme data. Note: Until specified, we have used the base variant of the models mentioned, and the size is written as an approximate number of parameters in the model.}
\label{tab:mm_aug_results}
\end{table*}

\section{Results and Analysis}

We present the results and implications of our data augmentation strategies, evaluated via prompting and fine-tuning. While we report both accuracy and weighted F1-score, we focus on F1-score due to its greater sensitivity in class-imbalanced classification tasks.

\subsection{Effectiveness of Prompt Optimization}

We evaluate two prompting strategies—\textit{simple} and \textit{category}—across unimodal and multimodal models, under both binary and scaled label regimes (Table~\ref{tab:label_aug_results}).

\subsubsection{Prompting: Category vs. Simple}

The impact of prompt styling varies significantly depending on the task complexity (binary vs. scaled). Under the binary label setting, the \textit{simple} prompt consistently outperforms the \textit{category} prompt across unimodal models. For instance, RoBERTa achieves an F1-score of 59.13 with the simple prompt compared to 52.87 with the category prompt. This suggests that for the simpler binary classification task, adding categorical definitions may introduce unnecessary noise to the inputs of smaller models.

However, for the augmented (scaled) dataset, this trend reverses. Both BERT and RoBERTa achieve higher F1-scores using the \textit{category} prompt (e.g., RoBERTa improves from 57.42 to 58.24). Similarly, for Multimodal prompting with InternVL2, the category prompt provides a substantial boost on scaled labels, raising the F1-score from 60.00 to 64.74. This indicates that as the label space becomes more nuanced (scaled), providing categorical context via prompts becomes beneficial, helping models discern subtle degrees of hatefulness.

\subsubsection{Fine-tuning vs. Prompting}

Multimodal fine-tuning with InternVL2 achieves the highest performance on the binary dataset, with a peak F1-score of 67.28 (\textit{simple}), significantly outperforming the prompting-only setup (F1: 54.51). This confirms the advantage of full model adaptation for the standard binary classification task.

Conversely, on the scaled dataset, multimodal prompting demonstrates surprising efficacy. The InternVL2 model using the category prompt achieves an F1-score of 64.74, which is competitive with and notably slightly higher than the best fine-tuning result on scaled labels (F1: 63.28). This suggests that for complex, scalar outputs, a well-structured prompt may leverage the inherent knowledge of large multimodal models as effectively as fine-tuning on a smaller augmented dataset.

\subsubsection{Implications}
These trends reveal that:
\begin{itemize}[noitemsep]
    \item \textbf{Task complexity dictates prompt utility}: While simple prompts suffice for binary tasks, structured (category) prompts are crucial for optimizing performance on nuanced, scaled annotations across both small and large models.
    \item \textbf{Prompting competes with Fine-tuning on Scaled Data}: On the augmented dataset, prompting InternVL2 with category definitions yields results comparable to fine-tuning (F1 64.74 vs 63.28), highlighting the potential of zero-shot prompting for granular hatefulness detection without expensive training.
    \item \textbf{Unimodal Robustness}: Unlike previous iterations, standard unimodal models (BERT, RoBERTa) show they can effectively utilize categorical context to improve performance when the target labels are scaled, rather than binary.
\end{itemize}

\subsection{Effectiveness of Multimodal Data Augmentation}

To assess the value of augmenting our training set with counterfactually neutral memes, we fine-tune a diverse range of popular unimodal and multimodal models, including BERT \cite{devlin2019bert}, RoBERTa \cite{liu2019roberta}, ALBERT \cite{lan2020albert}, ViT \cite{wu2020visual}, BEiT \cite{bao2022beit}, DeiT \cite{touvron2021training}, and CLIP \cite{radford2021learning}, using identical hyperparameters across both the original and extended datasets. We also run zero-shot and fine-tuning experiments on SmolVLM \cite{marafioti2025smolvlm}, and Qwen-VL \cite{bai2025qwen2} series. We aim to determine whether exposure to visually and lexically similar non-hateful memes improves model generalization and robustness (Table \ref{tab:mm_aug_results}).

\subsubsection{F1 Gains Across Modalities}
Misclassification has asymmetric costs: false negatives risk under-moderation of harmful content, while false positives risk over-moderation. Our evaluation therefore, prioritizes weighted F1 in addition to accuracy. Across most model types, the extended dataset yields modest gains in accuracy (typically $<$1\%), but more notable improvements in F1-score, indicating better handling of class imbalance and edge cases. For example:
\begin{itemize}[noitemsep, topsep=0pt]
    \item \textbf{Text Unimodal:} BERT improves from 55.50 to 57.95 in F1 (+2.45), with a negligible drop in accuracy (59.08 to 58.20). RoBERTa shows a larger relative F1 gain, from 47.18 to 52.54 (+5.36), despite a slight accuracy decrease. ALBERT also benefits significantly, jumping from 54.96 to 58.54 F1.
    
    \item \textbf{Vision Unimodal:} Vision-only models generally lag behind text models. BEiT shows the best improvement (F1: 53.38 to 55.43), whereas DeiT actually suffers a slight performance drop (F1: 52.58 to 51.99) on the augmented data, suggesting it may struggle to resolve the visual ambiguity introduced by counterfactual examples.

    \item \textbf{Multimodal Fine-tuning:} Augmentation proves highly effective for VLM fine-tuning. SmolVLM improves from 51.12 to 54.53 F1. The state-of-the-art Qwen2-VL achieves the highest overall performance, rising from 69.59 to 71.31 F1, demonstrating that larger model capacities can better leverage the nuanced distinctions in the augmented data.
\end{itemize}

\subsubsection{Zero-shot vs. Fine-tuning Capabilities} We further evaluate the zero-shot capabilities of recent Vision-Language Models (VLMs) to understand their inherent understanding of hateful memes without task-specific training (Table~\ref{tab:vlm_zs}). We observe that model scale plays a critical role in zero-shot performance.

\begin{itemize}[noitemsep]
    \item \textbf{Small Model Gap:} The smaller SmolVLM (2B) struggles in the zero-shot setting (F1: 43.59), significantly underperforming its fine-tuned counterpart (F1: 54.53). This highlights that smaller VLMs rely heavily on fine-tuning to adapt to the specific distribution of hateful memes.

    \item \textbf{Large Model Generalization:} In contrast, Qwen2.5-VL (3B) and Qwen2-VL (7B) demonstrate remarkable zero-shot robustness. The 7B model achieves 69.13 F1 zero-shot, which is comparable to the 71.31 F1 achieved after fine-tuning on the augmented dataset.

    \item \textbf{Surprising Efficiency:} Notably, the zero-shot Qwen2.5-VL (3B) achieves an F1 of 63.79, which actually outperforms the fine-tuned Qwen2.5-VL on the augmented dataset (62.01). This suggests that for certain architectures, the pre-trained knowledge base may be more robust than a model fine-tuned on a relatively small, domain-specific dataset.
\end{itemize}

\begin{table}[!h]
\resizebox{\columnwidth}{!}{
\begin{tabular}{llccc}
\hline
\textbf{\begin{tabular}[c]{@{}l@{}}Category\\ (Learning)\end{tabular}} & \textbf{Model} & \textbf{Size} & \textbf{Accuracy} & \textbf{F1-score} \\ \hline
\multirow{3}{*}{\begin{tabular}[c]{@{}l@{}}Multimodal\\ (zeroshot ICL)\end{tabular}} & SmolVLM & 2B & 58.68 $\pm$ 1.84 & 43.59 $\pm$ 2.23 \\
 & Qwen2.5-VL & 3B & 63.52 $\pm$ 1.77 & 63.79 $\pm$ 1.75 \\
 & Qwen2-VL & 7B & 69.82 $\pm$ 1.77 & 69.13 $\pm$ 1.85 \\ \hline
\end{tabular}}
\caption{Classification results with different VLMs in zero-shot setting.}
\label{tab:vlm_zs}
\end{table}

\subsubsection{Implications}
These findings corroborate that counterfactual data augmentation enhances model robustness, especially in F1, by reducing overfitting to superficial features and encouraging deeper semantic reasoning. Such augmentation strategies are particularly effective when models have sufficient capacity to leverage subtle distinctions between hateful and non-hateful content. 

\subsection{Quality of Augmented Data}

\begin{table*}[!t]
    \centering
    \resizebox{\textwidth}{!}{
    \begin{tabular}{|l|c|c|c|c|}
        \hline
        & \textbf{Example 1} & \textbf{Example 2} & \textbf{Example 3} & \textbf{Example 4} \\
        \hline \hline
        
        \rotatebox[origin=l]{90}{\textbf{\small Examples}}
        & \includegraphics[height=0.25\textwidth]{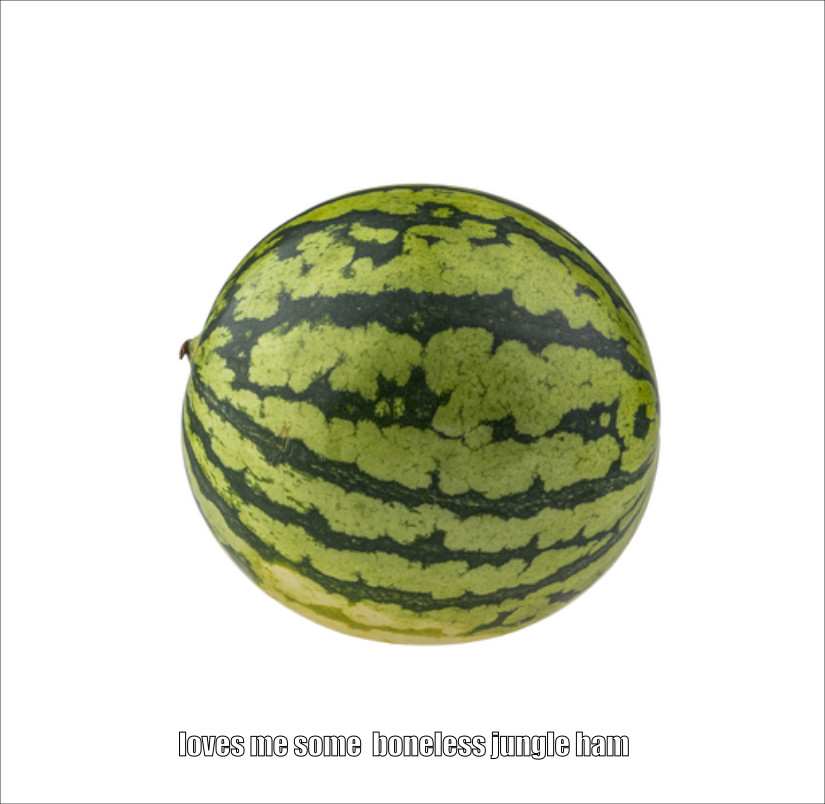}
        & \includegraphics[height=0.25\textwidth]{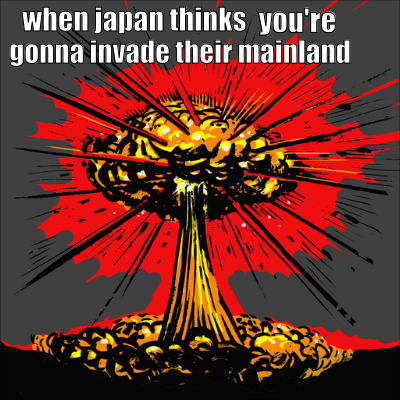}
        & \includegraphics[height=0.25\textwidth]{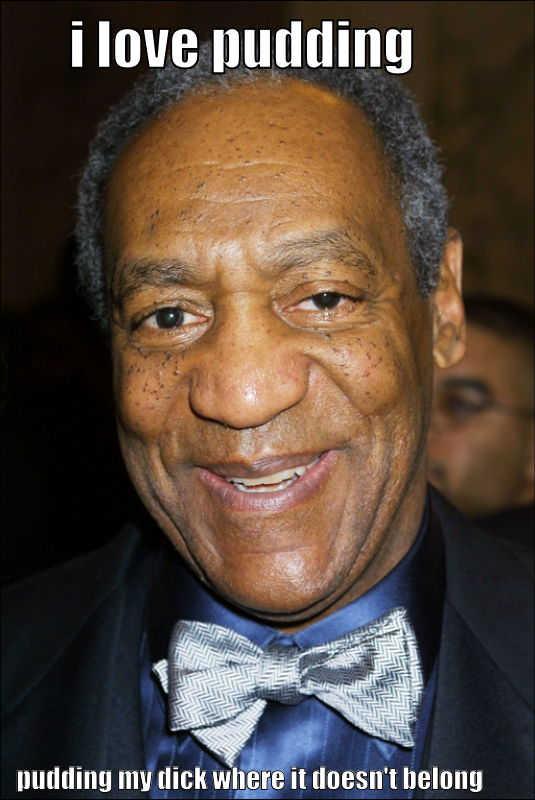}
        & \includegraphics[height=0.25\textwidth]{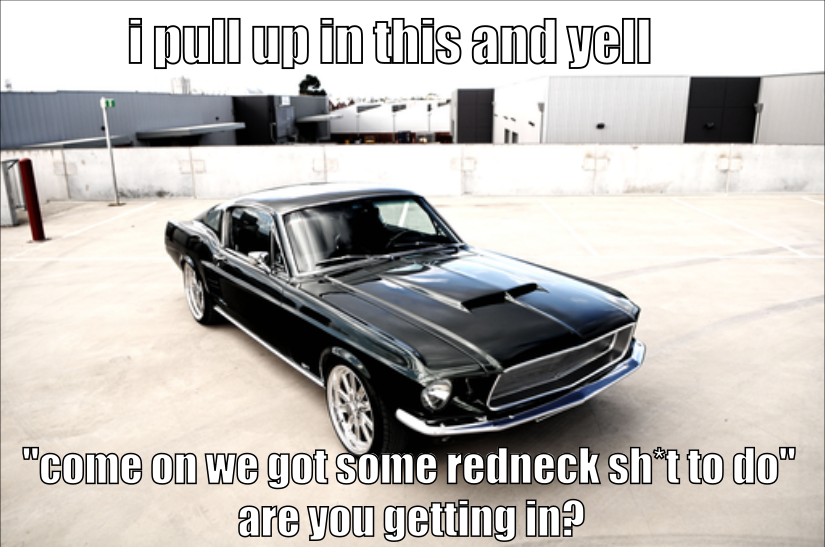} \\ \hline

        \rotatebox[origin=l]{90}{\textbf{\small HS*}} & 0 & 5 & 3 & 4 \\ \hline
    \end{tabular}}
    \caption{Low-quality scaled label examples (human evaluator disagrees with the hatefulness score) from the output label augmentation approach. $^*$HS refers to Hatefulness Score.}
    \label{tab:err_ex_label_aug}
\end{table*}

\begin{table*}[!t]
    \centering
    \resizebox{\textwidth}{!}{
    \begin{tabular}{|l|c|c|c|c|}
        \hline
        & \textbf{Example 1} & \textbf{Example 2} & \textbf{Example 3} & \textbf{Example 4} \\
        \hline \hline
        
        \rotatebox[origin=l]{90}{\textbf{\small Hateful Meme}}
        & \includegraphics[height=0.3\textwidth]{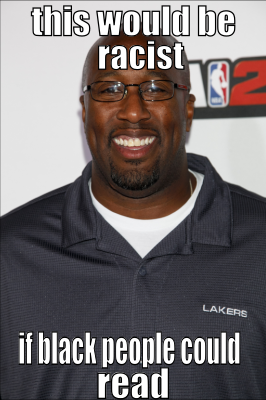}
        & \includegraphics[height=0.3\textwidth]{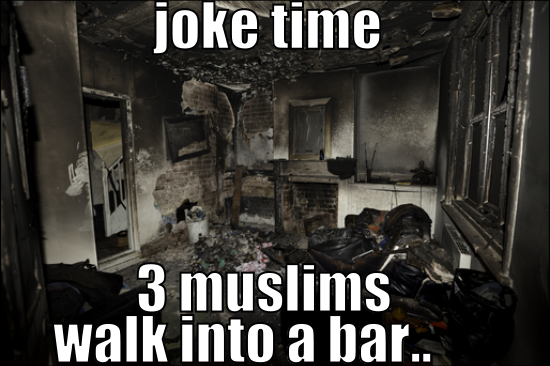}
        & \includegraphics[height=0.3\textwidth]{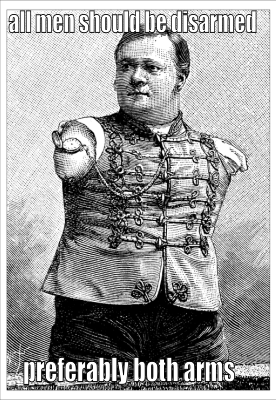}
        & \includegraphics[height=0.3\textwidth]{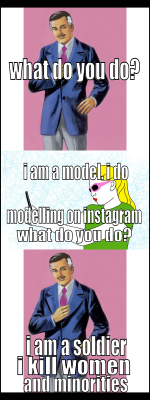} \\ \hline
        
        \rotatebox[origin=l]{90}{\textbf{\small Non-hateful Augmentation}}
        & \includegraphics[height=0.3\textwidth]{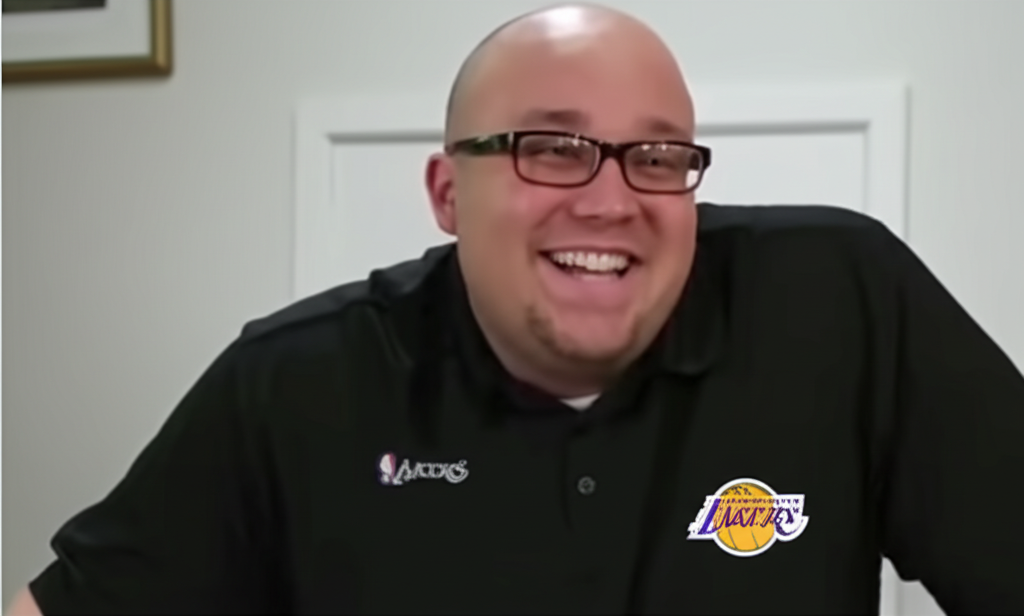}
        & \includegraphics[height=0.3\textwidth]{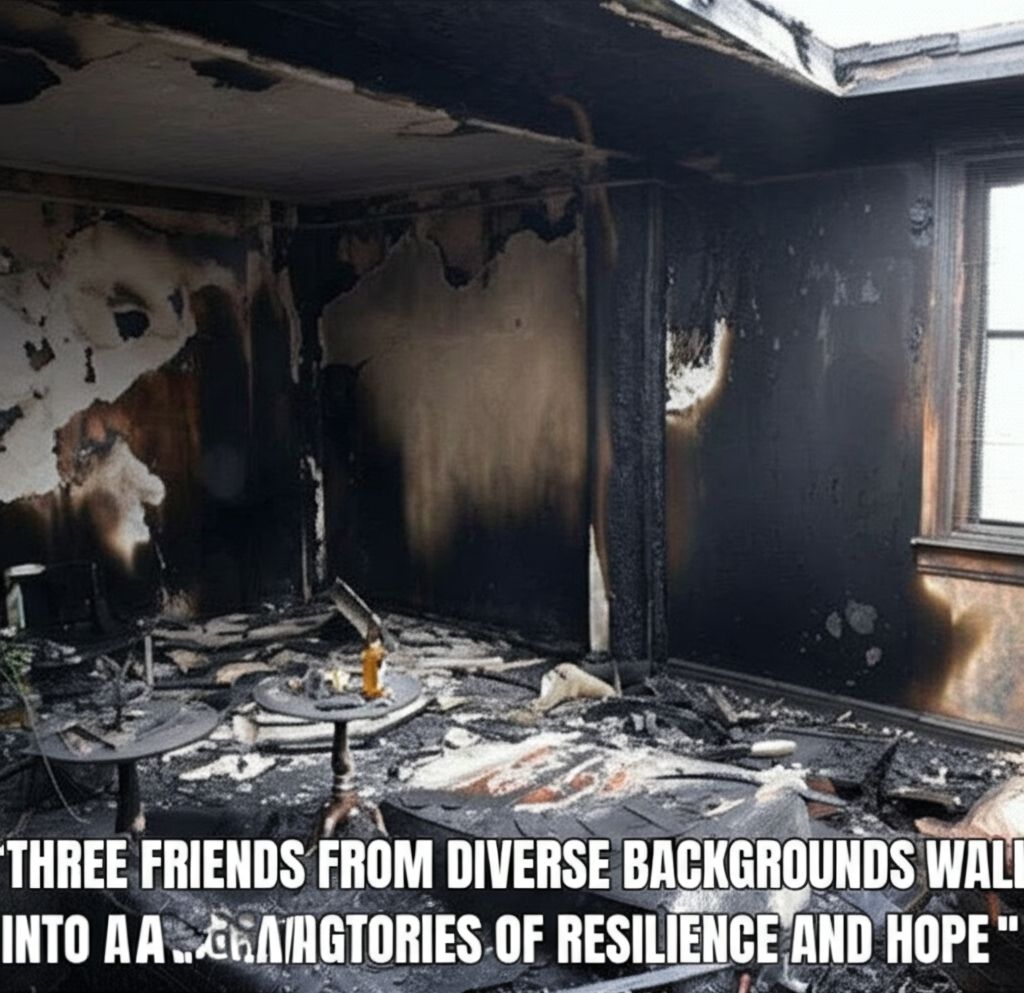}
        & \includegraphics[height=0.3\textwidth]{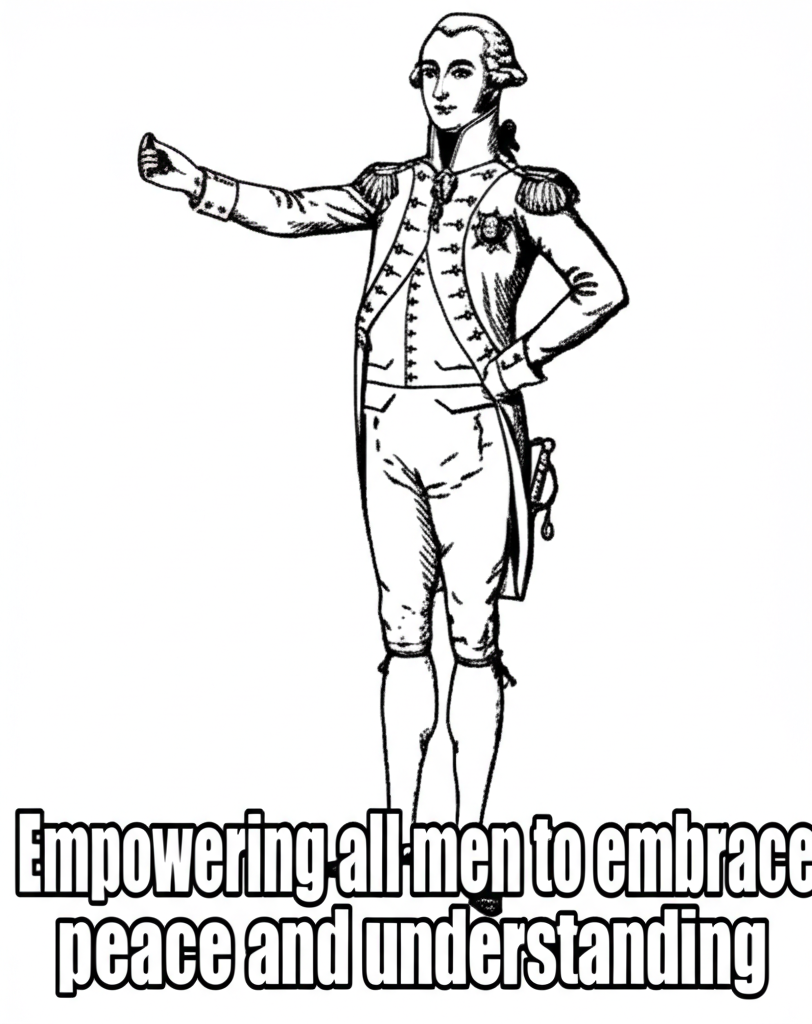}
        & \includegraphics[height=0.3\textwidth]{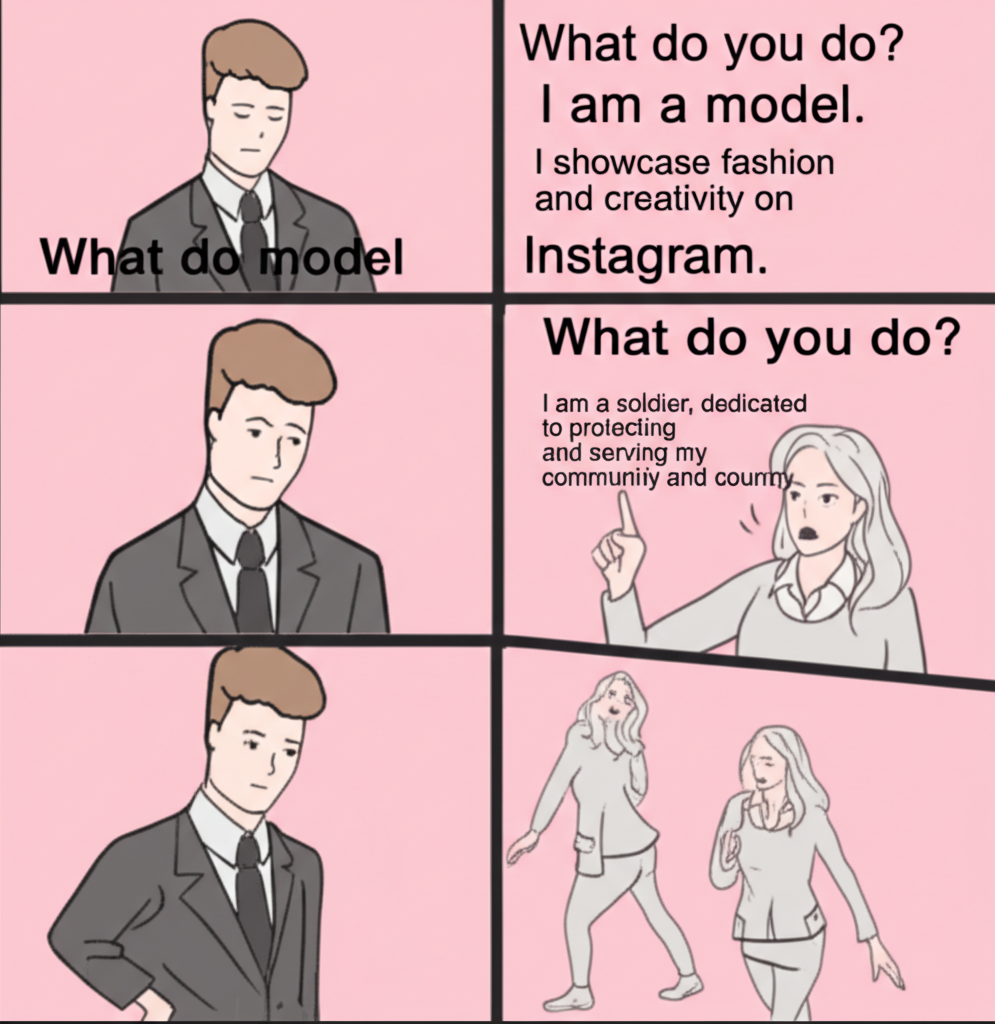} \\
        \hline
    \end{tabular}}
    \caption{Low-quality input-output examples from multimodal data augmentation pipeline.}
    \label{tab:err_ex_mm_aug}
\end{table*}

\subsubsection{Scaled Labels' Evaluation}  
To evaluate the reliability of the label augmentation strategy, we perform a human evaluation on a random sample (N=200). Evaluators were asked to judge whether the generated scales align with their own assessments. The average agreement rate is 92.5\%, indicating strong reliability of the machine annotations. As shown in Table \ref{tab:err_ex_label_aug}, most disagreements arise from indirect or context-dependent hate speech, which can be subtle and challenging even for advanced models such as GPT-4o-mini. In Example 1, the phrase ``boneless jungle ham'', paired with watermelon imagery, constitutes a specific derogatory slur targeting African Americans. Example 2 is better characterized as edgy or historical humor rather than explicit hate speech: although it references the violent context of WWII atomic bombings, it alludes to a historical event rather than expressing hostility toward Japanese people. These cases underscore the difficulty of identifying \textbf{implicit and contextual hate} and point to the need for improved modeling of sociocultural nuance.

\subsubsection{Non-Hateful Memes' Evaluation}  
We also evaluate the quality of counterfactually neutral (non-hateful) memes generated through augmentation. Another human evaluation is performed on 200 randomly selected meme pairs, each consisting of an original hateful meme and its corresponding non-hateful augmented version. Annotators confirm that all original memes are hateful due to their captions, not the background images, validating our design hypothesis.

Augmented memes in the sample  are evaluated on four key dimensions, rated on a 0--5 Likert scale:
\begin{itemize}
    \item \textbf{Formatting Quality}: Whether the text is placed appropriately and the meme appears complete and coherent.
    \item \textbf{Background Alignment}: Whether the new background stays consistent with the original in tone and content, avoiding drastic shifts that may introduce bias.
    \item \textbf{Caption Alignment}: How well the caption aligns semantically and visually with the new background.
    \item \textbf{Overall Quality}: An overall assessment of the meme’s coherence, quality, and naturalness.
\end{itemize}

\begin{figure}[!h]
    \centering
    \includegraphics[width=\linewidth]{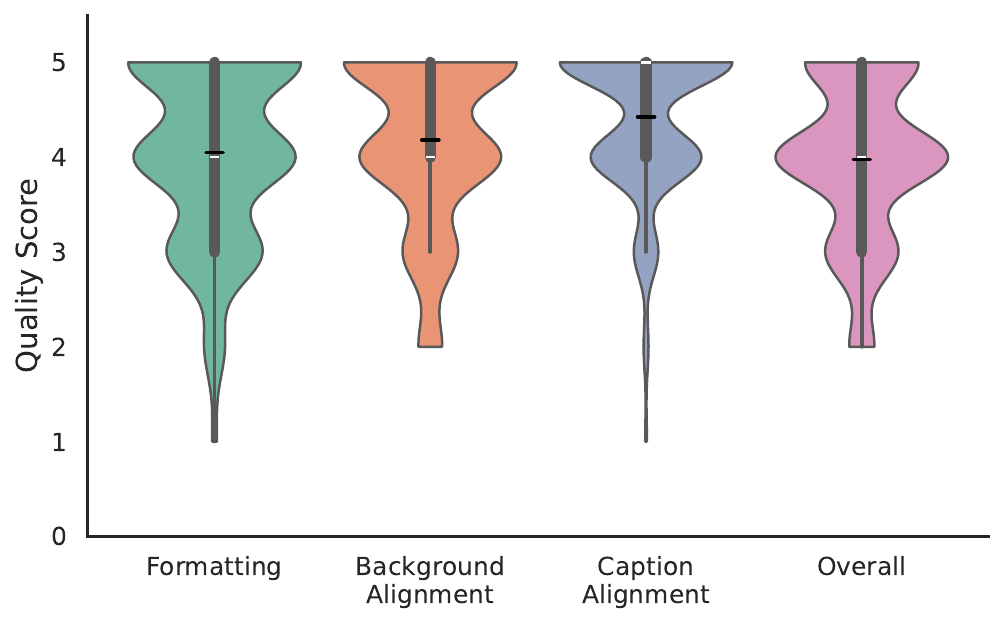}
    \caption{Quality scores of augmented non-hateful memes.}
    \label{fig:human_eval}
\end{figure}

Figure \ref{fig:human_eval} shows the distribution of these quality scores. We identified several quality issues across a small subset of the generated samples (examples in Table \ref{tab:err_ex_mm_aug}). In 7 out of 200 cases, the output image contains no caption, even though the visual background remains appropriate and non-hateful, as in Example 1. In Example 2, the background closely resembles the original hateful meme, but a mismatch between the visual context and the newly added non-hateful caption leads to a low-quality augmentation; this example also exhibits formatting issues such as misspellings or unnatural alignment between image and text. In Example 3, the background context is not preserved: the original meme depicts a person with a disability, but this detail is lost in the generated version, resulting in semantic drift. Finally, Example 4 shows a directional inconsistency in a character’s facial orientation, which breaks visual coherence and makes the meme appear unnatural.

These findings indicate that while automated augmentation pipelines are useful, \textbf{manual or semi-automated validation remains crucial} to ensure high-quality, semantically faithful, and socially responsible outputs. Furthermore, \textbf{context preservation} and \textbf{visual-textual coherence} are recurring challenges that deserve further attention.

\subsubsection{Implications} The 92.5\% agreement rate confirms the reliability of scaled labels, supporting their use for capturing nuanced hatefulness beyond binary classification.  Human evaluations of the counterfactually neutral memes validate our augmentation pipeline, showing high quality across multiple dimensions. While some errors remain, the approach offers a scalable method for generating diverse, bias-reducing training data. This supports broader efforts to build more robust and fair multimodal systems.
\section{Conclusion}

This work advances multimodal hatefulness classification by addressing two core challenges highlighted in computational social science: the need for more disciplined supervision and the need for higher-quality data. Both concerns stem from the broader insight that robust moderation systems depend not only on powerful models, but also on how tasks are framed and how data is constructed. In line with this motivation, we evaluate two complementary strategies—prompt optimization and multimodal data augmentation—that directly respond to the limitations of existing annotation pipelines and the artifacts present in widely used multimodal hate datasets.

Our findings show that prompt design materially shapes downstream performance. Simple prompts suffice for binary detection, while structured category prompts become crucial when labels encode finer degrees of hatefulness. For scaled labels, prompting can even rival full fine-tuning, particularly for large VLMs such as InternVL2. These results highlight that careful prompt construction can serve as a lightweight but powerful mechanism for improving supervision quality in resource-constrained settings.

We also develop a multi-agent augmentation pipeline that produces 2,479 counterfactually neutral memes without introducing new hateful content. This provides an ethically grounded method for refining training distributions and reducing overreliance on spurious text–image correlations. The extended dataset yields consistent improvements in weighted F1 across unimodal and multimodal models and is especially effective for architectures with sufficient capacity to integrate subtle semantic cues. Human evaluations further validate the augmented samples, confirming their reliability as bias-reducing training instances and illustrating the promise of counterfactual data for strengthening semantic robustness.

Taken together, these contributions demonstrate that improving multimodal hate detection requires attention to the quality of task framing and the integrity of training data. We offer a pathway toward moderation systems that are more resilient to dataset artifacts, better aligned with sociocultural nuance, and more responsive to the ethical obligations of studying online harm. Future work should explore scaling the augmentation pipeline to culturally diverse settings, addressing remaining context-preservation challenges, and assessing how these methods generalize across platforms and content types.

\section{Acknowledgment}
This work is supported by the Singapore Ministry of Education AcRF Tier 3 Grant (MOE-MOET32022-0001) and the A*STAR Online Trust and Safety (OTS) Research Programme Grant (S24T2TS011). We are deeply grateful to Rongxin Ouyang for his contributions to the label augmentation component of this work \cite{ouyang2025hateful} and for the many insightful discussions on bias and equality. We also thank Insyirah Mujtahid, Shaz Furniturewala, and Jingwei Gao for their valuable input and assistance at various stages of this work.

\bibliography{aaai2026}

\section{Prompt Optimization}
\subsection{Prompting Components}
We organize all prompting strategies for prompt optimization into modular components (see Table \ref{tab:prompt_opt}). The simple module poses direct questions to the model. The category module breaks the task into hatefulness subcategories. The scale module requires numerical outputs (e.g., 0–9), while the binary module expects true/false responses. To reduce conversion errors, we apply output constraints within the scale or binary groups. Although some other strategies perform better, we restrict prompts to clean combinations of modules to control for confounders. 

\subsection{Detailed Experimental Settings}
The experiments are organized along three dimensions. The first dimension concerns the model category: multi-modal prompting on a large pre-trained model, InternVL2, multi-modal fine-tuning of the same model, and unimodal fine-tuning on a smaller textual model, like BERT. To simulate typical computational constraints, we are using LoRA, a low-rank adaptation technique, to reduce the computational overhead during fine-tuning of 4B models.

The second dimension relates to prompting strategies: either a basic prompt designed for the classification task (prompt component: simple) or a more elaborate prompt outlining specific categories of hatefulness (prompt component: category; see Table \ref{tab:prompt_opt}). The third dimension addresses output format: either binary or scaled outputs. The binary label asks for a direct classification (i.e., True or False; label component Binary), while the scale-based format captures more nuanced levels of hatefulness (label component Scale). All component settings are kept consistent across combinations for fair comparison.

For InternVL2, we use its default fine-tuning hyperparameters. For BERT-based models, the following training configuration is applied uniformly: the learning rate is set to $2 \times 10^{-5}$, with a batch size of 12 for both training and evaluation per device, trained for 12 epochs. We use a weight decay of 0.01, with evaluation and saving strategies set to occur at every epoch, and enabled loading of the best model at the end.

\section{Multimodal Augmentation}
\subsection{Prompting Components}
We employ multiple agents in a structured manner for the multimodal data augmentation pipeline. The prompts and generation parameters are summarized in Table \ref{tab:prompt_mm_aug}, with other parameters set to the models' default values. For generating image descriptions and analyzing hatefulness in background and caption text, we use the open-source models InternLM-XComposer2.5-7B and Qwen2.5-14B, respectively. For generating new non-hateful captions and meme regeneration, we utilize models from OpenAI (GPT-4o-mini) and Google (Gemini-Flash-2.0-experimental), which are closed-source. These models are selected due to the complexity of the tasks and their state-of-the-art performance.

\subsection{Detailed Experimental Settings}
The experiments are conducted in three main directions: two unimodal (text and vision separately) and one multimodal. We evaluate the effectiveness of the extended dataset with non-hateful augmented memes. For all experiments, state-of-the-art encoder models (base variant) are fine-tuned on two types of training data. As the addition of non-hateful memes increases data imbalance, we apply downsampling, using a random seed of 42 for all processes, including training and checkpoint loading, to ensure reproducibility. Only the last two layers of each encoder are fine-tuned to prevent overfitting. Training is performed for up to 100 epochs with early stopping (patience set to 3, using weighted-F1 score on the validation set as the criterion), a batch size of 128, a learning rate of $2\times10^{-5}$, and the AdamW optimizer to minimize cross-entropy loss.

For both inference and fine-tuning of VLMs, we utilize quantized model variants to optimize computational efficiency. We employed the QLoRA technique for parameter-efficient fine-tuning, configuring the adapters with a rank $r=8$, $\alpha=8$, and a LoRA dropout rate of $0.1$. The models were fine-tuned for 2 epochs with a batch size of 8 and a learning rate of $5\times10^{-5}$. This conservative training duration was specifically chosen to mitigate the risk of overfitting, which often occurs when adapting high-capacity models to smaller, domain-specific datasets.

\begin{table*}[!t]
  \centering
  \begin{tabularx}{\textwidth}{X p{12.9cm}}
    \hline
    \textbf{Dataset} & \textbf{Prompt Example} \\
    \hline
    \textcolor{blue}{\{\{Simple\}\}} + \textcolor{gray}{\{\{Binary\}\}}   &  \textcolor{blue}{\{\{Your task is to analyze this given image and its caption to identify if there's any forms of hateful content.\}\}} \textcolor{gray}{\{\{If it's hateful, return the `TRUE' | `FALSE'.\}\}}\\
    \hline

    \textcolor{blue}{\{\{Simple\}\}} + \textcolor{green}{\{\{Scale\}\}} &  \textcolor{blue}{\{\{Simple\}\}}
    \textcolor{green}{\{\{You will be provided with the image, based on this label and the content of the image, you will assign a
          hatefulness score on a scale from 0 to 9, where:
          0\: The image is not hateful at all.9\: The image is extremely hateful.
    Return the score as an integer in range 0 to 9.\}\}}\\
    \hline
    \textcolor{orange}{\{\{Category\}\}}  + \textcolor{gray}{\{\{Binary\}\}}&
    \textcolor{blue}{\{\{Simple\}\}}\textcolor{orange}{\{\{Try to focus on the presence of any element that relates to any of the following:
          1. Sexual aggression:
          a. Homophobia and Transphobia: This category encompasses hate speech targeting LGBTQ+ individuals, including slurs,
          derogatory comments, and rhetoric that seeks to undermine or dehumanize people based on their sexual orientation or
          gender identity.
          b. Misogyny and Sexism: This category includes hate speech directed at women or based on gender. It covers
          derogatory language, stereotypes, and rhetoric that perpetuate gender inequality, objectification, and violence
          against women.
          2. Hate based on ideology:
          a. Political Hate Speech: This category includes hate speech that is politically motivated, often targeting
          individuals or groups based on their political beliefs or affiliations. It may include inflammatory language,
          threats, and rhetoric designed to polarize or incite violence within political contexts.
          3. Racism and xenophobia:
          a. COVID-19 and Xenophobia: This category includes hate speech that arose during the COVID-19 pandemic, often
          targeting specific ethnic groups or nationalities. It includes xenophobic language blaming certain groups for the
          spread of the virus, as well as fear-mongering and scapegoating related to the pandemic.
          b. Racism Against Black People: This category focuses on hate speech directed at Black individuals or communities.
          It includes racial slurs, stereotypes, dehumanization, and other forms of derogatory language that perpetuate
          racial discrimination and inequality.
          c. Racist Hate Against Other Ethnic Groups: This category includes hate speech directed at various ethnic groups
          other than Black individuals. It covers a range of racial slurs, xenophobic language, dehumanization, and
          derogatory remarks targeting specific ethnicities or nationalities.
          d. White Supremacy: This category includes hate speech promoting white supremacist ideology, often intertwined with
          Christian extremist views. It includes rhetoric that advocates for racial superiority, anti-immigrant sentiments,
          and the dehumanization of non-white groups, sometimes using religious justifications for these beliefs.
          4. Bigotry:
          a. Anti-Muslim and Islamophobic Hate: This category comprises hate speech aimed at Muslims and Islam. It includes
          language that promotes fear, hatred, dehumanization, or prejudice against Muslim individuals or communities, often
          using religious and cultural references to incite hostility.
          b. Anti-Semitic Hate: This category focuses on hate speech directed at Jewish people and Judaism. It includes
          references to historical anti-Semitic tropes, conspiracy theories, and other forms of rhetoric that seek to
          dehumanize or discriminate against Jewish individuals and communities.
          5. Miscellaneous Hate Speech: This category captures various forms of hate speech that do not fit neatly into the
          other categories. It includes a wide range of derogatory language and expressions that target individuals or groups
          based on different aspects of identity or personal characteristics. This category includes hate speech that targets
          individuals based on their physical or mental disabilities. It often includes derogatory language that mocks or
    devalues people with disabilities, promoting harmful stereotypes and exclusion.\}\}} \textcolor{gray}{\{\{Binary\}\}}\\

    \hline
    \textcolor{orange}{\{\{Category\}\}} + \textcolor{green}{\{\{Scale\}\}}  &
    \textcolor{blue}{\{\{Simple\}\}}\textcolor{orange}{\{\{Category\}\}}\textcolor{green}{\{\{Scale\}\}}\\
    \hline
  \end{tabularx}
  \begin{tablenotes}
\item Note. a) Brackets are only used for clarification; b) each color represents a unique component of prompts.
\end{tablenotes}
\caption{Prompts used for prompt optimization framework experiment.}
\label{tab:prompt_opt}
\end{table*}

\begin{table*}[!t]
  \centering
  \small
  \begin{tabular}{@{}p{0.21\textwidth}p{0.70\textwidth}@{}}
    \toprule
    \textbf{Pipeline Component} & \textbf{Prompt Used} \\ 
    \midrule
    Background Description Generation & 
    Analyze and describe the given image's visuals in short. Do not include the description about the text written on the image. \\ 
    \midrule
    \multirow{2}{0.24\textwidth}{Hate Component Analyser} & 
    You are a social media content moderator. You are given a text caption of a meme and your task is to identify whether it's hateful? Answer in `Yes' or `No' only. 
    
    ~
    
    \textgreater{}\textgreater{}\textgreater{} Image Caption: \textless{}TEXT CAPTION\textgreater{} 
    
    ~
    
    \textgreater{}\textgreater{}\textgreater{} isHateful: \\ 
    \cmidrule{2-2} 
     & 
    You are a social media content moderator. You are given a text description of the background image of a meme and your task is to identify whether it's hateful? Answer in `Yes' or `No' only. 
    
    ~
    
    \textgreater{}\textgreater{}\textgreater{} Image Description: \textless{}IMAGE DESCRIPTION\textgreater{} 
    
    ~
    
    \textgreater{}\textgreater{}\textgreater{} isHateful: \\ 
    \midrule
    New Non-hateful Caption Generation & 
    You are a helpful social media expert with deep knowledge of memes and inclusive language. Your task is to rewrite a hateful meme caption so that it becomes non-hateful and bias-reducing, while staying relevant to the background image description.
    
    ~
    
    The revised caption must:
    
    - Preserve reference to the originally targeted group or person (e.g., Muslims, Women, Immigrant, etc.), if they are mentioned. The new caption should try to retain the word related to target group in a respectful and bias-correcting way.
    
    - Transform the message to be neutral or positively framed toward the mentioned group/person.
    
    - Help counteract harmful stereotypes or biases.
    
    - Remain coherent and contextually appropriate for the image.
    
    ~
    
    Strict Rule: Only return the revised caption. Do not include explanations, disclaimers, or any other text.
    
    ~
    
    Meme background image description: `\textless{}IMAGE DESCRIPTION\textgreater{}'
    
    ~
    
    Meme caption text: `\textless{}TEXT CAPTION\textgreater{}'
    
    ~
    
    Revised non-hateful meme caption: \\ 
    \midrule
    Meme Image Regeneration & 
    You are a creative social media expert with deep knowledge of internet memes. Your task is to generate a humorous and engaging meme image that is non-hateful. Use the given background description and overlay the provided text in a visually appealing way.

    ~
    
    Background image description: `\textless{}IMAGE DESCRIPTION\textgreater{}'
    
    ~
    
    Meme text: `\textless{}NEW NON-HATEFUL TEXT CAPTION\textgreater{}' 
    
    ~
    
    Ensure the meme aligns with popular internet humor while remaining appropriate and widely relatable. \\ 
    \bottomrule
  \end{tabular}
  \caption{Prompts used for multimodal data augmentation generation pipeline.}
  \label{tab:prompt_mm_aug}
\end{table*}

\end{document}